\pdfoutput=1

\documentclass[11pt]{article}

\usepackage[]{emnlp2021}

\usepackage{times}
\usepackage{latexsym}

\usepackage[T1]{fontenc}

\usepackage[utf8]{inputenc}

\usepackage{microtype}

\usepackage{bm}
\usepackage{graphicx}
\usepackage{amsmath}
\usepackage{amsfonts}
\usepackage{linguex}
\usepackage{cleveref}
\usepackage{soul}
\usepackage{caption}
\usepackage{subcaption}
\usepackage{tabularx, colortbl}
\usepackage{booktabs}
\usepackage{arydshln}

\definecolor{blue-violet}{rgb}{0.54, 0.17, 0.89}
\definecolor{bluegray}{rgb}{0.4, 0.6, 0.8}
\definecolor{bleudefrance}{rgb}{0.19, 0.55, 0.91}
\definecolor{darkblue}{rgb}{0.0, 0.0, 0.55}
\definecolor{denim}{rgb}{0.08, 0.38, 0.74}
\definecolor{mediumblue}{rgb}{0.0, 0.0, 0.8}

\definecolor{carminered}{rgb}{1.0, 0.0, 0.22}
\definecolor{crimsonglory}{rgb}{0.75, 0.0, 0.2}

\definecolor{xgreen}{rgb}{0.88, 0.95, 0.83}

\newcommand{\hlc}[2][yellow]{{%
    \colorlet{foo}{#1}%
    \sethlcolor{foo}\hl{#2}}%
}

\title{\textsc{CLIFF}: Contrastive Learning for Improving Faithfulness and Factuality in Abstractive Summarization}

\author{Shuyang Cao \and Lu Wang \\
  Computer Science and Engineering \\
  University of Michigan \\
  Ann Arbor, MI \\
  \texttt{\{caoshuy, wangluxy\}@umich.edu} \\}

\begin{document}
\maketitle

\begin{abstract}

We study generating abstractive summaries that are faithful and factually consistent with the given articles. 
A novel contrastive learning formulation is presented, which leverages both reference summaries, as positive training data, and automatically generated erroneous summaries, as negative training data, to train summarization systems that are better at distinguishing between them. 
We further design four types of strategies for creating negative samples, to resemble errors made commonly by two state-of-the-art models, BART and PEGASUS, found in our new human annotations of summary errors.
Experiments on XSum and CNN/Daily Mail show that our contrastive learning framework is robust across datasets and models.
It consistently produces more factual summaries than strong comparisons with post error correction, entailment-based reranking, and unlikelihood training, according to QA-based factuality evaluation. 
Human judges echo the observation and find that our model summaries correct more errors. 

\end{abstract}
\section{Introduction}

Large pre-trained Transformers have yielded remarkable performance on abstractive summarization~\cite{liu-lapata-2019-text,lewis-etal-2020-bart,zhang2020pegasus} with impeccable fluency, yet their summaries often contain factually inconsistent content~\cite{maynez-etal-2020-faithfulness,zhang-etal-2020-optimizing,goyal-durrett-2020-evaluating}, even for state-of-the-art models.
Three types of remedies have been proposed: running a separately learned error correction component~\cite{dong-etal-2020-multi}, removing noisy training samples~\cite{nan-etal-2021-entity,goyal-durrett-2021-annotating}, and modifying the Transformer architecture~\cite{huang-etal-2020-knowledge,zhu-etal-2021-enhancing}. 
Yet they either rely on heuristically created data for error handling, falling short of generalization, or require learning a large number of new parameters, and summary informativeness is often sacrificed.

\begin{figure}[t]
    \centering
    \small
    \setlength{\tabcolsep}{3pt}
    \begin{tabular}{p{0.47\textwidth}}
    \toprule
        \textbf{XSum Article}: The Fermanagh MLA Phil Flanagan tweeted after Tom Elliott appeared on a BBC radio programme in May 2014. $\ldots$ 
        ``I wonder if he will reveal how many people he harassed and shot as a member of the UDR.''$\ldots$ \\
        \midrule
        \textbf{Contrastive learning} (our method): A Sinn Féin MLA has been ordered to apologise and pay compensation to a former member of the \hlc[yellow!70]{ Ulster Defence Regiment} (UDR). \\
        \midrule
        \textbf{Cross-entropy}: A Sinn Féin MLA has agreed to pay compensation to a former \hlc[cyan!40]{Ulster Unionist Party} (UDR) MP after he tweeted that he had harassed and shot people as a member of the party. \\
        \midrule
        \textbf{Entailment reranking}: A Sinn Féin MLA has agreed to pay compensation to a former \hlc[cyan!40]{Ulster Unionist Party} (UDR) councillor for a tweet he sent about him. \\
        \midrule
        \textbf{Unlikelihood}: An MLA has been ordered to apologise and pay compensation to a former \hlc[cyan!40]{loyalist} MP for a remark he made about him while serving in \hlc[cyan!40]{the Ministry of Defence}. \\
        
    \bottomrule
    \end{tabular}
    \caption{
    Sample article and system summaries by different methods. 
    Our contrastive learning model trained on low confidence system outputs correctly generates the \hlc[yellow!70]{full name}.
    Comparisons using cross-entropy loss,
    beam reranking by entailment scores~\cite{kryscinski-etal-2020-evaluating}, and unlikelihood objective~\cite{Welleck2020Neural} over negative samples all produce \hlc[cyan!40]{unfaithful content}.
    }
    \label{fig:intro_sample}
\end{figure}

Our goal is to train abstractive summarization systems that generate both faithful and informative summaries in an end-to-end fashion.
We observe that, while the commonly used maximum likelihood training optimizes over references, there is no guarantee for the model to distinguish references from incorrect generations~\cite{Holtzman2020The,Welleck2020Neural}. 
Therefore, potential solutions reside in designing new learning objectives that can effectively inform preferences of factual summaries over incorrect ones. 

Concretely, we hypothesize that including factually inconsistent summaries (i.e., \textit{negative samples}) for training, in addition to references (i.e., \textit{positive samples}), let models become better at differentiating these two types of summaries. 
Although using negative samples has been effective at text representation learning, e.g., word2vec~\cite{mikolov2013distributed} and BERT~\cite{devlin-etal-2019-bert}, there exist two major challenges for it to succeed in concrete language tasks. 
First, \textit{a suitable training objective} is critical to avoid performance degradation~\cite{saunshi2019theoretical}. 
Second, it is nontrivial to \textit{construct ``natural'' samples} that mimic the diverse errors made by state-of-the-art systems that vary in words and syntax~\cite{goyal-durrett-2021-annotating}. 

To address both challenges, we first propose a new framework, \textbf{\textsc{CLIFF}}, that uses \underline{c}ontrastive \underline{l}earning for \underline{i}mproving \underline{f}aithfulness and \underline{f}actuality of the generated summaries.\footnote{Our code and annotated data are available at \url{https://shuyangcao.github.io/projects/cliff_summ}.} 
Contrastive learning (CL) has obtained impressive results on many visual processing tasks, such as image classification~\cite{khosla2020supervised,pmlr-v119-chen20j} and synthesis~\cite{park2020contrastive,zhang2021cross}. 
Intuitively, CL improves representation learning by compacting positive samples while contrasting them with negative samples.
Here, we design a task-specific CL formulation that teaches a summarizer to expand the margin between factually consistent summaries and their incorrect peers.

Moreover, we design four types of strategies with different variants to construct negative samples by \textit{editing reference summaries} via rewriting entity-/relation-anchored text, and using \textit{system generated summaries} that may contain unfaithful errors. 
Importantly, these strategies are inspired by our new annotation study on  errors made by state-of-the-art summarizers---models fine-tuned from BART~\cite{lewis-etal-2020-bart} and PEGASUS~\cite{zhang2020pegasus}---on two benchmarks: XSum~\cite{narayan-etal-2018-dont} and CNN/DailyMail (CNN/DM)~\cite{NIPS2015_5945}. 

We fine-tune pre-trained large models with our contrastive learning objective on XSum and CNN/DM. 
Results based on QuestEval~\cite{scialom2020QuestEval}, a QA-based factuality metric of high correlation with human judgments,
show that our models trained with different types of negative samples uniformly outperform strong comparisons, including using a summarizer with post error correction and reranking beams based on entailment scores to the source.
Moreover, compared with unlikelihood training method that penalizes the same negative samples~\cite{Welleck2020Neural}, our summaries also obtain consistently better QuestEval scores. 
Human evaluation further confirms that our models consistently reduce both extrinsic and intrinsic errors over baseline across datasets. 

\section{Related Work}
\label{sec:related_work}

\paragraph{Factuality Improvement and Evaluation.} 
Neural abstractive summaries often contain unfaithful content with regard to the source~\cite{falke-etal-2019-ranking}. To improve summary factuality, three major types of approaches are proposed. First, a separate correction model is learned to fix errors made by the summarizers~\cite{zhao-etal-2020-reducing,chen-etal-2021-improving}, including replacing entities absent from the source~\cite{dong-etal-2020-multi} or revising all possible errors~\cite{cao-etal-2020-factual}. 
The second type targets at modifying the sequence-to-sequence architecture to incorporate relation triplets~\cite{cao2018faithful}, knowledge graphs~\cite{zhu-etal-2021-enhancing}, and topics~\cite{aralikatte-etal-2021-focus} to inform the summarizers of article facts. Yet additional engineering efforts and model retraining are often needed.  
Finally, discarding noisy samples from model training has also been investigated~\cite{nan-etal-2021-entity,goyal-durrett-2021-annotating}, however, it often leads to degraded summary informativeness. 
In comparison, our contrastive learning framework allows the model to be end-to-end trained and does not require model modification, thus providing a general solution for learning summarization systems.

Alongside improving factuality, we have also witnessed growing interests in automated factuality evaluation, since popular word-matching-based metrics, e.g., ROUGE, correlate poorly with human-rated factual consistency levels~\cite{gabriel-etal-2021-go,fabbri2021summeval}. 
Entailment-based scorers are designed at summary level~\cite{kryscinski-etal-2020-evaluating} and finer-grained dependency relation level~\cite{goyal-durrett-2020-evaluating}. 
QA models are employed to measure content consistency by reading the articles to answer questions generated from the summaries~\cite{wang-etal-2020-asking,durmus-etal-2020-feqa}, or considering the summaries for addressing questions derived from the source~\cite{scialom-etal-2019-answers}.
Though not focusing on evaluation, our work highlights that models can produce a significant amount of world knowledge which should be evaluated differently instead of as extrinsic hallucination~\cite{maynez-etal-2020-faithfulness}. We also show that world knowledge can possibly be distinguished from errors via model behavior understanding.

\medskip
\noindent \textbf{Training with negative samples} has been investigated in several classic NLP tasks, such as grammatical error detection~\cite{foster-andersen-2009-generrate} and dialogue systems~\cite{li-etal-2019-sampling}. 
Notably, negative sampling plays a key role in word representation learning~\cite{mikolov2013distributed} and training large masked language models, such as BERT and ALBERT, to induce better contextual representations~\cite{devlin-etal-2019-bert,Lan2020ALBERT:}. 
For text generation tasks, unlikelihood training is proposed to penalize the generation of negative tokens (e.g., repeated words) and sentences (e.g., contradictory responses in a dialogue system)~\cite{Welleck2020Neural,li-etal-2020-dont,he-glass-2020-negative}. 
We use contrastive learning that drives enhanced representation learning to better distinguish between factual and incorrect summaries, which encourages more faithful summary generation.

\paragraph{Contrastive Learning (CL) for NLP.}
CL has been a popular method for representation learning, especially for vision understanding~\cite{hjelm2018learning,pmlr-v119-chen20j}. 
Only recently has CL been used for training language models with self-supervision~\cite{fang2020cert}, learning sentence representations~\cite{gao2021simcse}, and improving document clustering~\cite{zhang2021supporting}. 
With a supervised setup, \citet{gunel2021supervised} adopt the contrastive objective to fine-tune pre-trained models on benchmark language understanding datasets. Using a similar idea, \citet{liu-liu-2021-simcls} enlarge the distances among summaries of different quality as measured by ROUGE scores. 

\section{\textsc{CLIFF}: Contrastive Learning Framework for Summarization}
\label{sec:method}

We design a contrastive learning (CL)-based training objective that drives the summarization model to learn a preference of faithful summaries over summaries with factual errors.
It is then used for fine-tuning BART~\cite{lewis-etal-2020-bart} and PEGASUS~\cite{zhang2020pegasus} for training summarization models.
Formally, let an article $x$ have a set of reference summaries $P$ (henceforth \textit{positive samples}) and another set of erroneous summaries $N$ (\textit{negative samples}). 
The contrastive learning objective is~\cite{khosla2020supervised,gunel2021supervised}:

\begin{equation}
    \fontsize{10}{11}\selectfont
    l_{cl}^{x} = - \frac{1}{\binom{|P|}{2}} \sum_{\substack{y_i, y_j \in P\\y_j \ne y_i}}
    \log \frac{\exp(\mathrm{sim} (\bm{h}_i, \bm{h}_j) / \tau)}{\sum\limits_{\substack{y_k \in P \cup N\\y_k \ne y_i}} \exp(\mathrm{sim} (\bm{h}_i, \bm{h}_k) / \tau)}
    \label{eq:contrast}
\end{equation}
where $\bm{h}_i$, $\bm{h}_j$, and $\bm{h}_k$ are representations for summaries $y_i$, $y_j$, and $y_k$.
$\mathrm{sim}(\cdot, \cdot)$ calculates the cosine similarity between summary representations. $\tau$ is a temperature and is set to $1.0$.

Importantly, summaries in $P$ and $N$ are included in the \textit{same batch} during training, so that the model acquires better representations to differentiate correct summaries from those with errors by comparing the two types of samples, thus maximizing the probabilities of the positive samples and minimizing the likelihoods of the corresponding negative samples.
The \textbf{CL objective} on the full training set, denoted as $\mathcal{L}_{CL}$, is the sum of losses $l_{cl}^{x}$ over all samples. 

To effectively employ CL in summarization, we need to address two challenges: (1) how to automatically construct both positive and negative samples, which are critical for CL efficacy~\cite{pmlr-v119-chen20j}, and (2) how to represent the summaries (i.e., $\bm{h}_{\ast}$).
Below we describe positive sample generation and options for $\bm{h}_{\ast}$, leaving the strategies for negative samples to \S~\ref{sec:sample_construct}.

\paragraph{Positive Sample Construction ($P$).}
Summarization datasets often contain a single reference for each article. To create multiple positive samples, in our pilot study,
we experiment with paraphrasing with synonym substitution~\cite{ren-etal-2019-generating}, randomly replacing words based on the prediction of masked language models~\cite{kobayashi-2018-contextual}, and back-translation~\cite{mallinson-etal-2017-paraphrasing}. We find back-translation to be best at preserving meaning and offering language variation, and thus use NLPAug\footnote{\url{https://github.com/makcedward/nlpaug}} to translate each reference to German and back to English. Together with the reference, the best translation is kept and added to $P$, if no new named entity is introduced. 

\paragraph{Summary Representation ($\bm{h}_{\ast}$).}
We use the outputs of the decoder's last layer, and investigate three options that average over \textit{all tokens}, \textit{named entity tokens}, and the \textit{last token} of the decoded summary. Entities and other parsing results are obtained by spaCy~\cite{spacy}. 
We further consider adding a multi-layer perceptron (MLP) with one hidden layer to calculate the final $\bm{h}_{\ast}$.

\medskip
\noindent \textbf{The final training objective} combines the typical cross-entropy loss $\mathcal{L}_{CE}$ and our contrastive learning objective: $\mathcal{L} = \mathcal{L}_{CE} + \lambda \mathcal{L}_{CL}$, where $\lambda$ is a scalar and set to $1.0$ for all experiments. 

\section{Summary Error Annotation and Model Behavior Analysis}
\label{sec:error_annotation}

We first describe annotating unfaithfulness errors by state-of-the-arts, i.e., models fine-tuned from BART and PEGASUS 
on XSum and CNN/DM.
We then probe into the model generation behavior that is indicative of errors, which guides the design of negative sample construction strategies.

600 ($150\times2\times2$) summaries are annotated to demonstrate how often do the models ``hallucinate", i.e., generating content not grounded by the source. 
To characterize errors, we annotate \textit{text spans} in summaries with 
(i) \textbf{intrinsic errors} caused by misconstructing phrases or clauses from the source; 
and (ii) \textbf{extrinsic errors} which include words not in the source that are either unverifiable or cannot be verified by Wikipedia. 
Content not covered by the article but can be validated by Wikipedia is annotated as \textbf{world knowledge}, and the models' behavior pattern when generating them differs from when they generate errors.

Two fluent English speakers with extensive experience in summary evaluation and error labeling are hired. For each sample, they are shown the article and two system summaries, and instructed to annotate text spans with the aforementioned errors and world knowledge. After labeling every 50 samples, the annotators discuss and resolve any disagreement. 
The Fleiss's Kappas on XSum and CNN/DM are $0.35$ and $0.45$. 

\begin{figure}[t]
    \centering
    \includegraphics[width=0.45\textwidth]{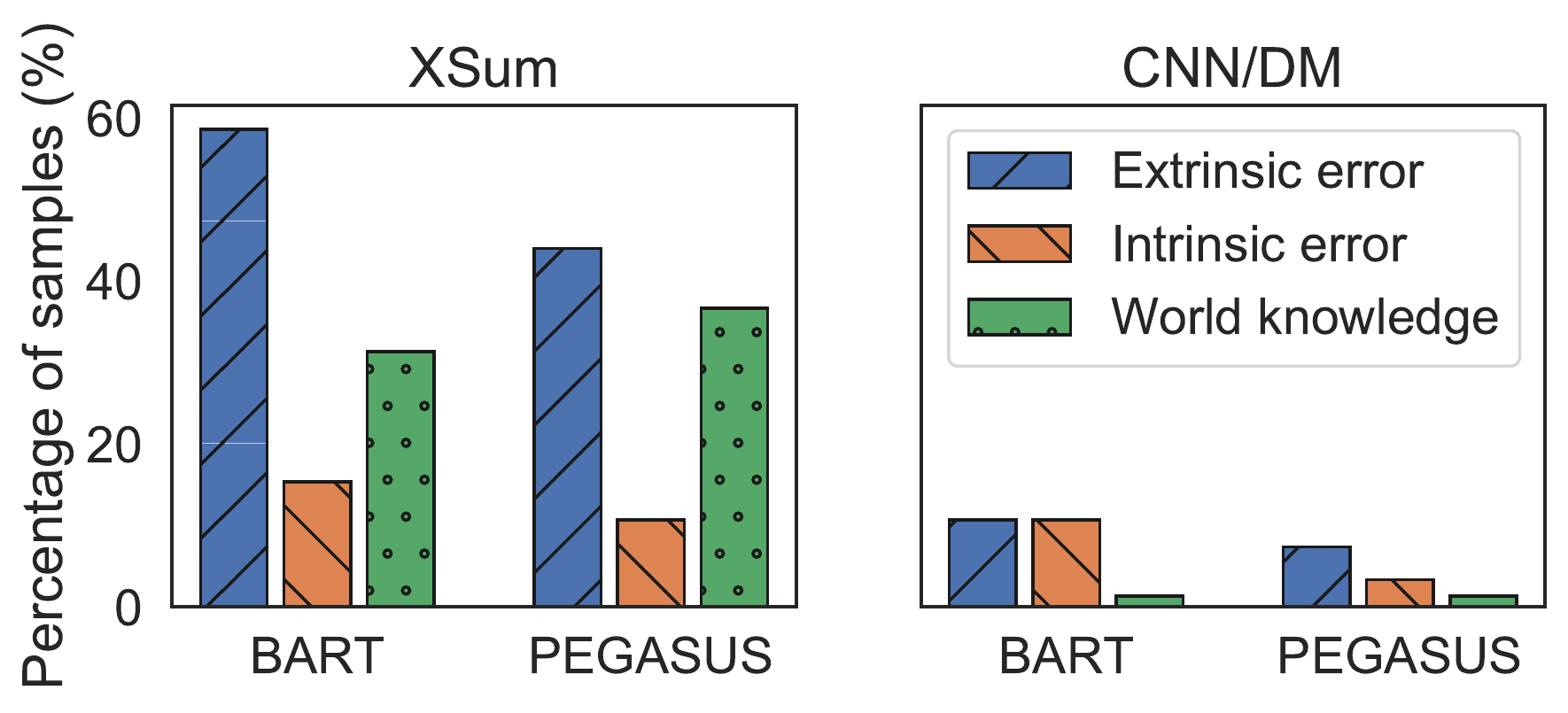}
    \caption{Percentage of samples with intrinsic and extrinsic error spans for models fine-tuned from BART and PEGASUS on XSum and CNN/DM.}
    \label{fig:error_type_dist}
\end{figure}

\medskip
\noindent \textbf{Error statistics} are displayed in Fig.~\ref{fig:error_type_dist}. 
Extrinsic errors dominate both datasets, especially on XSum. $58.7\%$ of summaries by BART (and $44.0\%$ by PEGASUS) contain at least one extrinsic error.
Noticeably, PEGASUS is a newer model pre-trained with a larger amount of data, thus contains less errors than BART and other older models studied for error annotations by~\newcite{maynez-etal-2020-faithfulness}. This observation also highlights the usage of our annotations for future development and evaluation of summarization models.

\begin{figure}[t]
    \centering
    \includegraphics[width=0.46\textwidth]{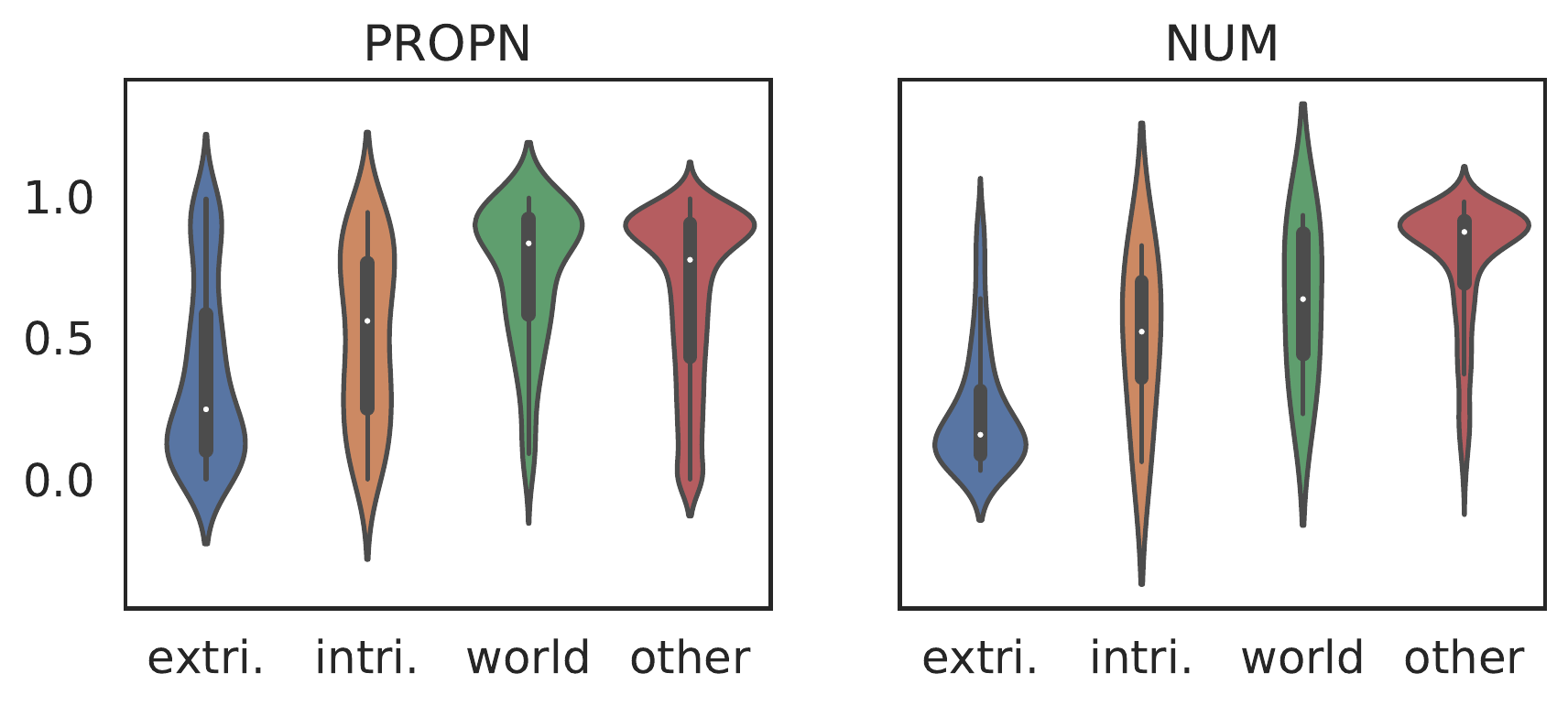}
    \caption{
    Probability distributions of generating the first tokens of proper nouns and numbers, grouped by extrinsic errors, intrinsic errors, world knowledge, and other correct tokens.}
    \label{fig:output_prob_propnum}
\end{figure}

\paragraph{Low confidence generation is indicative of extrinsic errors.}
Inspired by recent work that studies model prediction confidence~\cite{liu2021tokenlevel}, we examine generation probabilities for tokens of \textit{different part-of-speech (POS) tags}.
Fig.~\ref{fig:output_prob_propnum} shows salient results on the generation probabilities of the first token of a proper noun or a number (with additional analysis provided in Appendix~\ref{appendix:error_annotation}). 
As observed, model confidence tends to be lower for the first tokens of proper nouns and numbers if they are part of spans with \textit{extrinsic errors}.
Also note that world knowledge, which cannot be inferred from the source either, often has higher generation probability than extrinsic errors. 
Take this snippet generated by a fine-tuned BART as an example: \textit{``Manchester United captain Wayne Rooney's testimonial game against Manchester City$\ldots$''}. \textit{``Manchester City"} is an extrinsic error and \textit{``Wayne"} is produced as world knowledge. The model assigns a low probability of $0.10$ to the first token of \textit{``Manchester City''} and a high probability of $0.92$ to token \textit{``Wayne''}. 
This implies that model confidence can be a useful indicator for negative sample collection.

\section{Negative Sample Construction}
\label{sec:sample_construct}

Here we describe four strategies for constructing negative samples that 
\textit{modify the references} (\S~\ref{subsec:swapping}-\ref{subsec:conditional}) or use \textit{system generated summaries} (\ref{subsec:modelgeneration}). 

\begin{table}[t]
    
    \fontsize{8.5}{9}\selectfont
    \setlength{\tabcolsep}{2pt}
    \hspace{-1mm}
    \begin{tabular}{p{76mm}}
    \toprule
        \textbf{\textsc{Reference}}: A ``rare'' short-eared owl found emaciated in Flintshire is now recuperating well, the RSPCA have said. \\
        \midrule \midrule
        \textbf{\textsc{SwapEnt}}: Flintshire $\rightarrow$ Bettisfield \\
        $\Rightarrow$ A ``rare'' short-eared owl found emaciated in \textcolor{red!80!black}{\textbf{Bettisfield}} is now recuperating well, the RSPCA have said. \\
        \midrule
        \textbf{\textsc{MaskEnt}}:  A ``rare'' short-eared owl found emaciated in \texttt{[MASK]} is now recuperating well, the RSPCA have said. \\
        $\Rightarrow$ A ``rare'' short-eared owl found emaciated in a field in \textcolor{red!80!black}{\textbf{South Yorkshire}} is now recuperating well, the RSPCA have said. \\
        \midrule
        \textbf{\textsc{MaskRel}}:  A ``rare'' short-eared owl found \texttt{[MASK]} in \texttt{[MASK]} is now recuperating well, the RSPCA have said. \\
        $\Rightarrow$ A ``rare'' short-eared owl found \textcolor{red!80!black}{\textbf{dead} in \textbf{London}} is now recuperating well, the RSPCA have said. \\
        \midrule
        \textbf{\textsc{RegenEnt}}: A ``rare'' short-eared owl found emaciated in \rule{6mm}{0.15mm} \\
        $\Rightarrow$ A ``rare'' short-eared owl found emaciated in \textcolor{red!80!black}{\textbf{Nottinghamshire}} \textbf{is now at a wildlife centre to recover.} \\
        \midrule
        \textbf{\textsc{RegenRel}}:  A ``rare'' short-eared owl found \rule{6mm}{0.15mm} \\
        $\Rightarrow$ A ``rare'' short-eared owl found \textbf{in the grounds of a former coal mine is being cared for \textcolor{red!80!black}{by the RSPCA in Somerset.}} \\
        \midrule
        \textbf{\textsc{SysLowCon}}:  An injured \textcolor{red!80!black}{golden} owl found in a former coal mine in \textcolor{red!80!black}{Lancashire} is being cared for \textcolor{red!80!black}{by the RSPCA}. \\
    \bottomrule
    \end{tabular}
    \caption{
    Negative sample construction strategies (\S~\ref{sec:sample_construct}). For summaries edited from the reference, their differences are \textbf{bolded}. Introduced errors are in \textcolor{red!80!black}{\textbf{red}}. Text before \rule{6mm}{0.15mm} is the prefix for regeneration.
    }
    \label{tab:construction_example}
\end{table}

\subsection{Entity Swap}
\label{subsec:swapping}

Entity swap imitates intrinsic errors, as over $55\%$ of intrinsic errors in our annotations are found to contain named entities.
We construct negative samples by swapping named entities in the references with other randomly selected entities of the same entity type in the source (\textbf{\textsc{SwapEnt}}). One sample is constructed for each entity in the reference. 
Though this idea has been studied by \citet{kryscinski-etal-2020-evaluating}, they allow entities of different types to be used, e.g., a PERSON can be replaced by a LOCATION. 
Examples are displayed in Table~\ref{tab:construction_example}. 

\textsc{SwapEnt} has the advantage of not depending on any trained model. Yet it only introduces intrinsic errors and lacks the coverage for extrinsic errors, which is addressed by the following generation-based methods.

\subsection{Mask-and-fill with BART}
\label{subsec:unconditional}
To simulate extrinsic errors, we leverage large unconditional language models' capability of converting a sequence with masked tokens into a fluent and appropriate sequence. Specifically, we replace each named entity in a reference with a \texttt{[MASK]} token and encode it with BART (without any fine-tuning). BART then fills this partially masked summary with newly generated entities (\textbf{\textsc{MaskEnt}}). BART is chosen since it can fill \texttt{[MASK]} with varying number of tokens.
For each entity in the reference, we sample three summaries and only retain the ones containing at least one entity that is absent from both the source and the reference.

Up to now, the two introduced strategies both focus on incorrect named entities. To cover more diverse extrinsic and intrinsic errors~\cite{goyal-durrett-2020-evaluating}, we extend \textsc{MaskEnt} to contain relations (\textbf{\textsc{MaskRel}}). 
We first obtain dependency relations using Stanza~\cite{qi-etal-2020-stanza}, with each relation denoted as $<$\texttt{gov}, \texttt{rel}, \texttt{dep}$>$. 
To incorporate more context, we consider noun phrase spans enclosing the token of \texttt{gov} or \texttt{dep} if it is a content word and the noun phrase contains a named entity. 
Similar to \textsc{MaskEnt}, three negative samples are generated by BART based on the input with both \texttt{gov} and \texttt{dep} spans masked in the reference. Only the samples that introduce any new dependency relation that is not contained in the source nor the reference are kept. Specifically, we consider a match of a dependency relation as the same form or synonyms of its \texttt{gov} and and \texttt{dep} is found in the source or the reference with the same relation.

Both \textsc{MaskEnt} and \textsc{MaskRel} can create more extrinsic errors compared to other strategies introduced in this section, since negative samples are generated without being grounded on the source articles. However, their constructed negative samples may contained drifted topics that can be easily detected by a summarization model, resulting with less efficient training signals.

\subsection{Source-conditioned Regeneration}
\label{subsec:conditional}

To ground negative sample generation with the article, we further design a regeneration strategy based on conditional generation. For each named entity in the reference, we treat the text before it as a \textit{prompt}.
A summarizer, e.g., fine-tuned from BART or PEGASUS, first reads in the source using the encoder, then receives the prompt as the first part of the decoder output, and finally decodes the rest of the content based on nucleus sampling~\cite{Holtzman2020The} with a cumulative probability threshold of $0.7$. 
The prompt and the regenerated text comprise the final negative sample. This method is denoted as \textbf{\textsc{RegenEnt}}. 

We also extend entities to relations with expanded governor and dependent spans (\textbf{\textsc{RegenRel}}). Here, we consider a prompt as the text before the \texttt{gov} or \texttt{dep} span, whichever occurs first.  
For both \textsc{RegenEnt} and \textsc{RegenRel}, three negative samples are generated for each prompt, and a sample is kept if it introduces any new entity (for \textsc{RegenEnt}) or dependency relation (for \textsc{RegenRel}) with regard to the source and the reference. 

Negative samples generated by both methods are more relevant to the article than the mask-and-fill strategy, yet they may still miss certain types of errors and differ from real model outputs, since they are modified from the reference summaries.

\subsection{System Generation}
\label{subsec:modelgeneration}

Motivated by the model confidence analysis in \S~\ref{sec:error_annotation}, we explore using system generated summaries as negative samples. We first run fine-tuned BART or PEGASUS on the same training set to decode summaries. For each summary, we check the model confidence on the first token of each proper noun and number span. If the probability is below a threshold, we keep it as a negative sample (\textbf{\textsc{SysLowCon}}). The threshold is tuned by maximizing F1 based on our error annotations. 

We consider all beams at the last decoding step as candidates. We use beam sizes of $6$ and $4$ for XSum and CNN/DM.
Statistics of negative samples constructed by different strategies are in Appendix~\ref{appendix:dataset_stat}.

\section{Experiment Setup}
\label{sec:exp_setup}

\paragraph{Evaluation Metrics.} 
QuestEval~\cite{scialom2020QuestEval} is used as the main metric to evaluate summaries' factual consistency. 
Given an article and a summary, QuestEval first generates natural language questions for entities and nouns from both.  
A QA model then consumes the article to answer questions derived from the summary, producing a score. Another score is obtained from a QA model addressing article-based questions after reading the summary.
The final QuestEval score is the harmonic mean of the two. We use the version with learned weights for questions, which has shown high correlation with human judged consistency and relevance.

We further use FactCC~\cite{kryscinski-etal-2020-evaluating}, trained based on their negative sample construction method, to measure if the summary can be entailed by the source.
We also report ROUGE-L~\cite{lin-2004-rouge}. 
Both FactCC and ROUGE-L reasonably correlate with summary factuality as judged by human~\cite{pagnoni2021understanding}. 

Based on our error annotations, we report the correlations between each metric and the error rate---percentage of tokens being part of an error span, and the raw number of errors (Table~\ref{tab:metric_correlation}).
QuestEval correlates better on both aspects than other metrics.

\begin{table}[t]
    \centering
    \small
    \setlength{\tabcolsep}{4.5pt}
    \begin{tabular}{lcccc}
    \toprule
        \textbf{Metric} & \multicolumn{2}{c}{\textbf{XSum}} & \multicolumn{2}{c}{\textbf{CNN/DM}} \\
        & \textbf{\% of Err} & \textbf{\# of Err} & \textbf{\% of Err} & \textbf{\# of Err} \\
        \midrule
        QuestEval & \textbf{-0.43}\rlap{$^\ast$} & \textbf{-0.25}\rlap{$^\ast$} & \textbf{-0.33}\rlap{$^\ast$} & \textbf{-0.29}\rlap{$^\ast$} \\
        FactCC & -0.02 & -0.15\rlap{$^\ast$} & -0.13\rlap{$^\ast$} & -0.12\rlap{$^\ast$} \\
        ROUGE-1 & -0.16\rlap{$^\ast$} & -0.02 & -0.03 & -0.06 \\
        ROUGE-2 & -0.11\rlap{$^\ast$} & -0.05 & -0.02 & -0.04 \\
        ROUGE-L & -0.13\rlap{$^\ast$} & -0.03 & -0.06 & -0.08 \\
        \bottomrule
    \end{tabular}
    \caption{
    Pearson correlation between metrics and error rates and numbers of errors. 
    $\ast$: p-value $< 0.05$.
    }
    \label{tab:metric_correlation}
\end{table}

\paragraph{Comparisons.}
In addition to the models fine-tuned with cross-entropy loss (\textbf{\textsc{CrsEntropy}}), we consider reranking beams based on FactCC score (also our metric) at the last decoding step (\textbf{\textsc{EntailRank}}). 
We also include three common methods of improving factuality: 
(1) (\textbf{\textsc{Correction}}) fine-tunes BART to fix summary errors as a separate step~\cite{cao-etal-2020-factual}. 
(2) \textbf{\textsc{SubsetFT}} fine-tunes large models based on training samples without any dependency relation error~\cite{goyal-durrett-2021-annotating}, with released checkpoint only available for XSum. 
(3) \textbf{\textsc{FASum}} modifies Transformer by incorporating knowledge graphs for factual consistency~\cite{zhu-etal-2021-enhancing}, with model outputs only on CNN/DM.

Moreover, we compare with \textbf{unlikelihood training} that penalizes the probabilities of all tokens in a negative sample~\cite{li-etal-2020-dont}. Given a negative sample $y'$, the loss is defined as $-\sum_{t=1}^{| y' |} \log (1 - p(y'_t | y'_{1:t-1}, x))$, where $p(y'_t | y'_{1:t-1}, x)$ is the output probability at the $t$-th step. We combine the unlikelihood training objective with cross-entropy loss with equal weights for fine-tuning.

Lastly, we compare our negative sample strategies with negative samples constructed for training the FactCC scorer, denoted as \textbf{\textsc{FCSample}}. 
For CL only, we compare with using other samples in the same batch as negative samples (\textbf{\textsc{Batch}}), a common practice for CL-based representation learning~\cite{gao2021simcse, zhang2021supporting}.

\begin{table}[!t]
    \centering
    \small
    \setlength{\tabcolsep}{2.5pt}
    \begin{tabular}{lcccccc}
    \toprule
        \textbf{Model} & \multicolumn{3}{c}{\textbf{XSum}} & \multicolumn{3}{c}{\textbf{CNN/DM}} \\
        \cmidrule(lr){2-4} \cmidrule(lr){5-7}
         & \textbf{QEval} & \textbf{FC} & \textbf{R-L} & \textbf{QEval} & \textbf{FC} & \textbf{R-L} \\
         \midrule
         \multicolumn{7}{l}{\textit{Comparisons without Negative Samples}}\\
        \textsc{CrsEntropy} & 33.09 & 23.92 & 37.14 & 50.94 & 49.07 & 40.82 \\
        \textsc{EntailRank} & 32.95 & \textbf{38.45} & 36.55 & 51.02 & 49.84 & 40.89 \\
        \textsc{Correction} & 33.12 & 24.14 & 37.11 & 50.93 & 49.06 & 40.82 \\
        \textsc{SubsetFT} & 32.25 & 21.83 & 30.35 & - & - & - \\
        \textsc{FASum} & - & - & - & 50.73 &  50.58 & 37.18 \\
        \midrule
        \midrule
        \multicolumn{7}{l}{\textit{Comparisons with Unlikelihood Training}} \\
        \textsc{FCSample} & 32.93 & 24.46 & 33.96 & 50.60 & 35.09 & \hlc[xgreen]{41.22} \\
        \textsc{SwapEnt} & 32.90 & 23.94 & 34.96 & 49.77 & 32.37 & 40.18 \\
        \textsc{MaskEnt} & 33.21 & 24.22 & 33.89 & \hlc[xgreen]{51.01} & 48.57 & \hlc[xgreen]{\textbf{41.23}} \\
        \textsc{MaskRel} & 33.18 & 23.50 & 34.56 & 51.00 & 48.35 & \hlc[xgreen]{41.15}  \\
        \textsc{RegenEnt} & 32.41 & 24.12 & \hlc[xgreen]{37.08} & 50.97 & 48.59 & 41.07 \\
        \textsc{RegenRel} & 30.86 & 24.58 & \hlc[xgreen]{\textbf{37.18}} & \hlc[xgreen]{50.97} & 48.42 & \hlc[xgreen]{41.14} \\
        \textsc{SysLowCon} & 32.01 & \hlc[xgreen]{26.30} & 32.04 & 50.82 & 48.66 & 40.81 \\
        \midrule
        \multicolumn{7}{l}{\textit{Our Method:} \textsc{CLIFF}} \\
        \textsc{Batch} & 33.18  & 24.88 & 36.76 & 50.99 & \textbf{52.18} & 40.98 \\
        \hdashline
        \textsc{FCSample} & \hlc[xgreen]{33.15} & \hlc[xgreen]{24.50} & \hlc[xgreen]{36.72} & \hlc[xgreen]{51.02} & \hlc[xgreen]{49.62} & 41.06 \\
        \textsc{SwapEnt} & \hlc[xgreen]{33.30} & \hlc[xgreen]{25.67}\rlap{$^\ast$} & \hlc[xgreen]{35.60} & \hlc[xgreen]{\textbf{51.05}} & \hlc[xgreen]{50.96} & \hlc[xgreen]{40.89} \\
        \textsc{MaskEnt} & \hlc[xgreen]{33.32}\rlap{$^\ast$} & \hlc[xgreen]{25.73}\rlap{$^\ast$} & \hlc[xgreen]{36.02} & 50.98 & \hlc[xgreen]{49.04} & 41.06 \\
        \textsc{MaskRel} & \hlc[xgreen]{\textbf{33.35}}\rlap{$^\ast$} & \hlc[xgreen]{25.69}\rlap{$^\ast$} & \hlc[xgreen]{35.86} & \hlc[xgreen]{51.03} & \hlc[xgreen]{49.89} & 41.04 \\
        \textsc{RegenEnt} & \hlc[xgreen]{33.15} & \hlc[xgreen]{24.64} & 36.32 & \hlc[xgreen]{51.04} & \hlc[xgreen]{49.91} & \hlc[xgreen]{41.11} \\
        \textsc{RegenRel} & \hlc[xgreen]{33.22} & \hlc[xgreen]{25.39} & 36.21 & 50.96 & \hlc[xgreen]{49.48} & 41.11 \\
        \textsc{SysLowCon} & \hlc[xgreen]{\textbf{33.35}}\rlap{$^\ast$} & 25.47\rlap{$^\ast$} & \hlc[xgreen]{36.19} & \hlc[xgreen]{\textbf{51.05}} & \hlc[xgreen]{50.05} & \hlc[xgreen]{41.01} \\
        \bottomrule
    \end{tabular}
    \caption{
    Results of models fine-tuned from BART on XSum and CNN/DM. 
    QEval: QuestEval; FC: FactCC; R-L: ROUGE-L. 
    The best result per metric per dataset is \textbf{bolded}. 
    For models of unlikelihood training and CLIFF that use the same negative samples, the better of the two is highlighted with \hlc[xgreen]{green}. 
    $\ast$: our model is significantly better than \textsc{CrsEntropy} (approximation randomization test, $p < 0.005$).}
    \label{tab:main_result}
\end{table}

\section{Results}
\label{sec:results}

\subsection{Automatic Evaluation}
\label{subsec:autoeval}
We report results by models fine-tuned from BART and PEGASUS with different objectives and negative samples on XSum and CNN/DM in Tables~\ref{tab:main_result} and~\ref{tab:pegasus_result}. 
\textsc{CLIFF} models use a summary representation of averaging over all tokens with MLP projection, with other variants discussed in \S~\ref{subsec:ablaton}. Unless explicitly stated, comparison models are fine-tuned from the same large model used by \textsc{CLIFF}. 

First, comparing with other factuality improvement models (top of the tables), \textit{almost all \textsc{CLIFF} models trained with different negative samples uniformly produce higher QuestEval scores across datasets with both large models}, with the improvements more pronounced on XSum.
Importantly, ROUGE scores for \textsc{CLIFF} models are comparable or better than baselines trained with cross-entropy, e.g., on CNN/DM as in Table~\ref{tab:main_result}. 
A similar trend is observed with the FactCC metric, especially when using PEGASUS as the base model (Table~\ref{tab:pegasus_result}). 
Note that \textsc{EntailRank} tends to yield significantly higher FactCC scores, though it obtains lower QuestEval scores than the cross-entropy baseline. Human inspection finds that \textsc{EntailRank} can pick up beams with peculiar words of high FactCC scores, without improving factuality. 
Moreover, other comparisons based on post \textsc{Correction} and model engineering (\textsc{FASum}) only offer incremental gains. 
The sample selection-based method, \textsc{SubsetFT}, sacrifices ROUGE scores significantly. 
Overall, \textsc{CLIFF} demonstrates stronger generalizability.

\begin{table}[!t]
    \centering
    \small
    \setlength{\tabcolsep}{2.5pt}
    \begin{tabular}{lcccccc}
    \toprule
        \textbf{Model} & \multicolumn{3}{c}{\textbf{XSum}} & \multicolumn{3}{c}{\textbf{CNN/DM}} \\
        \cmidrule(lr){2-4} \cmidrule(lr){5-7}
         & \textbf{QEval} & \textbf{FC} & \textbf{R-L} & \textbf{QEval} & \textbf{FC} & \textbf{R-L} \\
         \midrule
         \multicolumn{7}{l}{\textit{Comparisons without Negative Samples}}\\
        \textsc{CrsEntropy} & 32.50 & 25.48 & \textbf{39.07} & 50.21 & 44.44 & 40.39 \\
        \textsc{EntailRank} & 32.42 & \textbf{41.90} & 38.47 & 50.15 & \textbf{61.04} & \textbf{40.67} \\
        \textsc{Correction} & 32.55 & 25.15 & 39.02 & 49.48 & 32.96 & 39.79 \\
        \midrule
        \midrule
        \multicolumn{7}{l}{\textit{Comparisons with Unlikelihood Training}} \\
        \textsc{FCSample} & 32.79 & \hlc[xgreen]{25.37} & 38.46 & 50.63 & 45.45 & 39.28 \\
        \textsc{SwapEnt} & 32.88 & 24.76 & 37.91 & 50.43 & 43.02 & \hlc[xgreen]{38.96} \\
        \textsc{MaskEnt} & 33.04 & \hlc[xgreen]{26.30} & 37.51 & 51.11 & 52.19 & \hlc[xgreen]{39.34} \\
        \textsc{MaskRel} & \hlc[xgreen]{33.08} & 24.38 & 38.05 & 51.14 & 52.93 & 39.31  \\
        \textsc{RegenEnt} & 32.89 & 24.46 & \hlc[xgreen]{38.47} & \hlc[xgreen]{51.11} & \hlc[xgreen]{52.90} & 39.23 \\
        \textsc{RegenRel} & 32.91 & 24.80 & \hlc[xgreen]{38.46} & 51.07 & \hlc[xgreen]{53.68} & \hlc[xgreen]{39.45} \\
        \textsc{SysLowCon} & 31.66 & \hlc[xgreen]{26.06} & 34.03 & \hlc[xgreen]{50.92} & 51.08 & 39.19 \\
        \midrule
        \multicolumn{7}{l}{\textit{Our Method:} \textsc{CLIFF}} \\
        \textsc{Batch} & 32.64  & 24.96 & 38.42 & 51.03\rlap{$^\ast$} & 51.81\rlap{$^\ast$} & 39.38 \\
        \hdashline
        \textsc{FCSample} & \hlc[xgreen]{32.96}\rlap{$^\ast$} & 25.28 & \hlc[xgreen]{38.58} & \hlc[xgreen]{51.00}\rlap{$^\ast$} & \hlc[xgreen]{51.80}\rlap{$^\ast$} & \hlc[xgreen]{39.37} \\
        \textsc{SwapEnt} & \hlc[xgreen]{33.09}\rlap{$^\ast$} & \hlc[xgreen]{25.09} & \hlc[xgreen]{38.58} & \hlc[xgreen]{51.16}\rlap{$^\ast$} & \hlc[xgreen]{52.97}\rlap{$^\ast$} & 38.95 \\
        \textsc{MaskEnt} & \hlc[xgreen]{33.09}\rlap{$^\ast$} & 25.75 & \hlc[xgreen]{38.12} & \hlc[xgreen]{51.13}\rlap{$^\ast$} & \hlc[xgreen]{53.60}\rlap{$^\ast$} & 39.24 \\
        \textsc{MaskRel} & 33.06\rlap{$^\ast$} & \hlc[xgreen]{25.28} & \hlc[xgreen]{38.37} & \hlc[xgreen]{\textbf{51.17}}\rlap{$^\ast$} & \hlc[xgreen]{53.34}\rlap{$^\ast$} & \hlc[xgreen]{39.36} \\
        \textsc{RegenEnt} & \hlc[xgreen]{33.09}\rlap{$^\ast$} & \hlc[xgreen]{24.48} & 38.33 & 50.99\rlap{$^\ast$} & 52.18\rlap{$^\ast$} & \hlc[xgreen]{39.28} \\
        \textsc{RegenRel} & \hlc[xgreen]{33.16}\rlap{$^\ast$} & \hlc[xgreen]{24.82} & 38.30 & \hlc[xgreen]{51.16}\rlap{$^\ast$} & 53.21\rlap{$^\ast$} & 39.25 \\
        \textsc{SysLowCon} & \hlc[xgreen]{\textbf{33.21}}\rlap{$^\ast$} & 25.18 & \hlc[xgreen]{38.18} & 50.85\rlap{$^\ast$} & \hlc[xgreen]{53.73}\rlap{$^\ast$} & \hlc[xgreen]{39.30} \\
        \bottomrule
    \end{tabular}
    \caption{
    Results of models fine-tuned from PEGASUS on XSum and CNN/DM. We report results on $5,000$ randomly selected samples on CNN/DM, due to long running time of QuestEval. 
    For models of unlikelihood training and CLIFF that use the same negative samples, the better of the two is highlighted with \hlc[xgreen]{green}. 
    $\ast$: our model is significantly better than \textsc{CrsEntropy} (approximation randomization test, $p < 0.005$).}
    \label{tab:pegasus_result}
\end{table}

Second, \textit{\textsc{CLIFF} is more effective and robust than unlikelihood training} with the same negative samples. 
According to Table~\ref{tab:main_result}, using 7 negative sample construction strategies on two datasets, \textsc{CLIFF} obtains higher QuestEval scores than unlikelihood training in 12 out of the 14 comparisons. Using PEGASUS, \textsc{CLIFF} also outperforms in 11 setups as listed in Table~\ref{tab:pegasus_result}. 
Similar trends are found on FactCC and ROUGE-L.
Another noteworthy piece is that \textsc{CLIFF}'s improvements over the cross-entropy baseline are more consistent, whereas unlikelihood training occasionally hurts factuality or ROUGE scores significantly. 
We believe the key advantage of \textsc{CLIFF} resides in its measure of representation similarities between positive and negative samples in the same batch, allowing models to better differentiate between correct and erroneous summaries.

Finally, among all variants, \textit{\textsc{CLIFF} trained with low confidence summaries as negative samples obtains the best QuestEval scores on the more abstractive dataset}. As seen in Table~\ref{tab:main_result}, using low confidence summaries also improves FactCC scores on both datasets, and enhances ROUGE-L on the more extractive dataset CNN/DM. 
This indicates that \textit{system generated summaries contribute more diverse errors made by existing models organically}, which are particularly suitable for our CL framework. As we use summaries generated by the same model for \textsc{CLIFF} training, one future direction is to use outputs by different models. 
For our mask-and-fill and source-conditioned regeneration strategies, we find that relation-anchored construction often beats their entity-anchored counterparts. This calls for efforts that steer the entity-driven methods to a more relation-focused direction.

\paragraph{Combining Strategies.}
We further show results by fine-tuning BARTs using samples based on combined negative sample construction strategies in Table~\ref{tab:combine_strategy}. As can be seen, \textit{combining \textsc{SysLowCon} and other strategies yields better QuestEval scores than models trained with negative samples by any single strategy}, except for \textsc{MaskEnt} and \textsc{RegenEnt} on XSum. This signifies the importance of covering diverse types of errors in negative samples.

\begin{table}[t]
    \centering
    \small
    \setlength{\tabcolsep}{2.5pt}
    \begin{tabular}{lcccccc}
    \toprule
        \textbf{Strategy} & \multicolumn{3}{c}{\textbf{XSum}} & \multicolumn{3}{c}{\textbf{CNN/DM}} \\
        \cmidrule(lr){2-4} \cmidrule(lr){5-7} 
        & \textbf{QEval} & \textbf{FC} & \textbf{R-L} & \textbf{QEval} & \textbf{FC} & \textbf{R-L} \\
        \midrule
        \textsc{SysLowCon} & 33.35 & 25.47 & \textbf{36.19} & 51.05 & 50.05 & \textbf{41.01} \\
        + \textsc{SwapEnt} & \textbf{33.40} & \textbf{25.50} & 35.50 & \textbf{51.32} & \textbf{53.95} & 40.57 \\
        + \textsc{MaskEnt} & 33.21 & 25.47 & 35.91 & 51.16 & 51.90 & 40.66 \\
        + \textsc{MaskRel} & 33.39 & 25.20 & 35.70 & 51.24 & 52.48 & 40.80 \\
        + \textsc{RegenEnt} & 33.31 & 25.07 & 35.94 & 51.21 & 51.86 & 40.91 \\
        + \textsc{RegenRel} & 33.38 & 24.97 & 36.03 & 51.13 & 50.85 & 40.97 \\
        \bottomrule
    \end{tabular}
    \caption{Results of fine-tuned BART with combinations of negative sample construction strategies.}
    \label{tab:combine_strategy}
\end{table}

\subsection{Human Evaluation}
\label{subsec:humaneval}

\begin{table}[t]
    \begin{subtable}[h]{0.48\textwidth}
    \small
    \setlength{\tabcolsep}{2pt}
        \begin{tabular}{lcccccc}
        \toprule
         & \multicolumn{3}{c}{\textbf{Inform.}} & \multicolumn{3}{c}{\textbf{Factual.}} \\
        \textbf{Model} & \textbf{Win}$\uparrow$ & \textbf{Tie} & \textbf{Lose}$\downarrow$ & \textbf{Win}$\uparrow$ & \textbf{Tie} & \textbf{Lose}$\downarrow$ \\
        \midrule
        \textsc{EntailRank} & 3.3 & 84.3 & \textbf{12.3} & 23.7 & 71.7 & \textbf{4.7} \\
        \textsc{ULL. MaskEnt} & 6.3 & 80.7 & 13.0 & 26.3 & 62.0 & 11.7 \\
        \textsc{CL. Batch} & 5.3 & 80.0 & 14.7 & 21.7 & 68.0 & 10.3 \\
        \textsc{CL. SysLowCon} & \textbf{8.7} & 78.3 & 13.0 & \textbf{31.3} & 61.7 & 7.0 \\
        \bottomrule
        \end{tabular}
        \vspace{-1mm}
        \caption{XSum}
        
    \end{subtable}
    
    \begin{subtable}[h]{0.48\textwidth}
    \small
    \setlength{\tabcolsep}{2pt}
        \begin{tabular}{lcccccc}
        \toprule
        & \multicolumn{3}{c}{\textbf{Inform.}} & \multicolumn{3}{c}{\textbf{Factual.}} \\
        \textbf{Model} & \textbf{Win}$\uparrow$ & \textbf{Tie} & \textbf{Lose}$\downarrow$ & \textbf{Win}$\uparrow$ & \textbf{Tie} & \textbf{Lose}$\downarrow$ \\
        \midrule
        \textsc{EntailRank} & 2.3 & 86.3 & 11.3 & 4.7 & 94.7 & \textbf{0.7} \\
        \textsc{ULL. MaskEnt} & \textbf{18.0} & 71.0 & 11.0 & 17.3 & 79.7 & 3.0 \\
        \textsc{CL. Batch} & 17.3 & 74.7 & \textbf{8.0} & \textbf{20.7} & 77.0 & 2.3 \\
        \textsc{CL. SysLowCon} & 15.7 & 75.7 & 8.7 & 20.0 & 77.7 & 2.3 \\
        \bottomrule
        \end{tabular}
        \vspace{-1mm}
        \caption{CNN/DM}
    \end{subtable}
    \caption{
    Percentages of summaries that are better than, tied with, or worse than \textsc{CrsEntropy}, in informativeness (Inform.) and factual consistency (Factual.) 
    The Krippendorff's $\alpha$s are $0.33$ and $0.62$ for the two aspects on XSum, and $0.34$ and $0.89$ on CNN/DM. Our CL method using low confidence summaries is more frequently rated as better for informativeness and factuality on the more abstractive dataset XSum. 
    }
    \label{tab:human_eval}
\end{table}

\paragraph{Pairwise Comparison with Cross-entropy.} 
We recruit the two human annotators for our summary error study, as well as another experienced annotator, to evaluate summary \textbf{informativeness} and \textbf{factual consistency}. For each article, the judges are shown summaries generated by the \textsc{CrsEntropy} model and four other systems. They then rate each system summary against the \textsc{CrsEntropy} summary.
All four summaries generated by different factuality-improved models are shown in random order without system names shown, ensuring the fair comparison among them.

We randomly pick $100$ articles from each dataset used in our error analysis study in \S~\ref{sec:error_annotation}, and evaluate summaries generated by \textsc{EntailRank}, unlikelihood training (\textsc{ULL}) with negative samples constructed by \textsc{MaskEnt}, and \textsc{CLIFF} models trained with \textsc{Batch} and \textsc{SysLowCon} negative samples. All are fine-tuned from BART. Detailed evaluation guidelines are in Appendix~\ref{appendix:human_eval}.

Table~\ref{tab:human_eval} shows that on the more abstractive XSum data \textit{CL trained with low confidence samples are more frequently rated as being more informative and more factual} than \textsc{CrsEntropy} summaries. This echos our automatic evaluations with QuestEval in \S~\ref{subsec:autoeval}.
On CNN/DM, all models trained with negative samples produce summaries with better informativeness and faithfulness.
In contrast, \textsc{EntailRank} summaries are less distinguishable from outputs by \textsc{CrsEntropy} on both datasets, as more ties are found. 
%
We show sample outputs in Fig.~\ref{fig:intro_sample}, with additional examples in Appendix~\ref{appendix:outputs}.

\begin{figure}[t]
    \centering
    \includegraphics[width=0.47\textwidth]{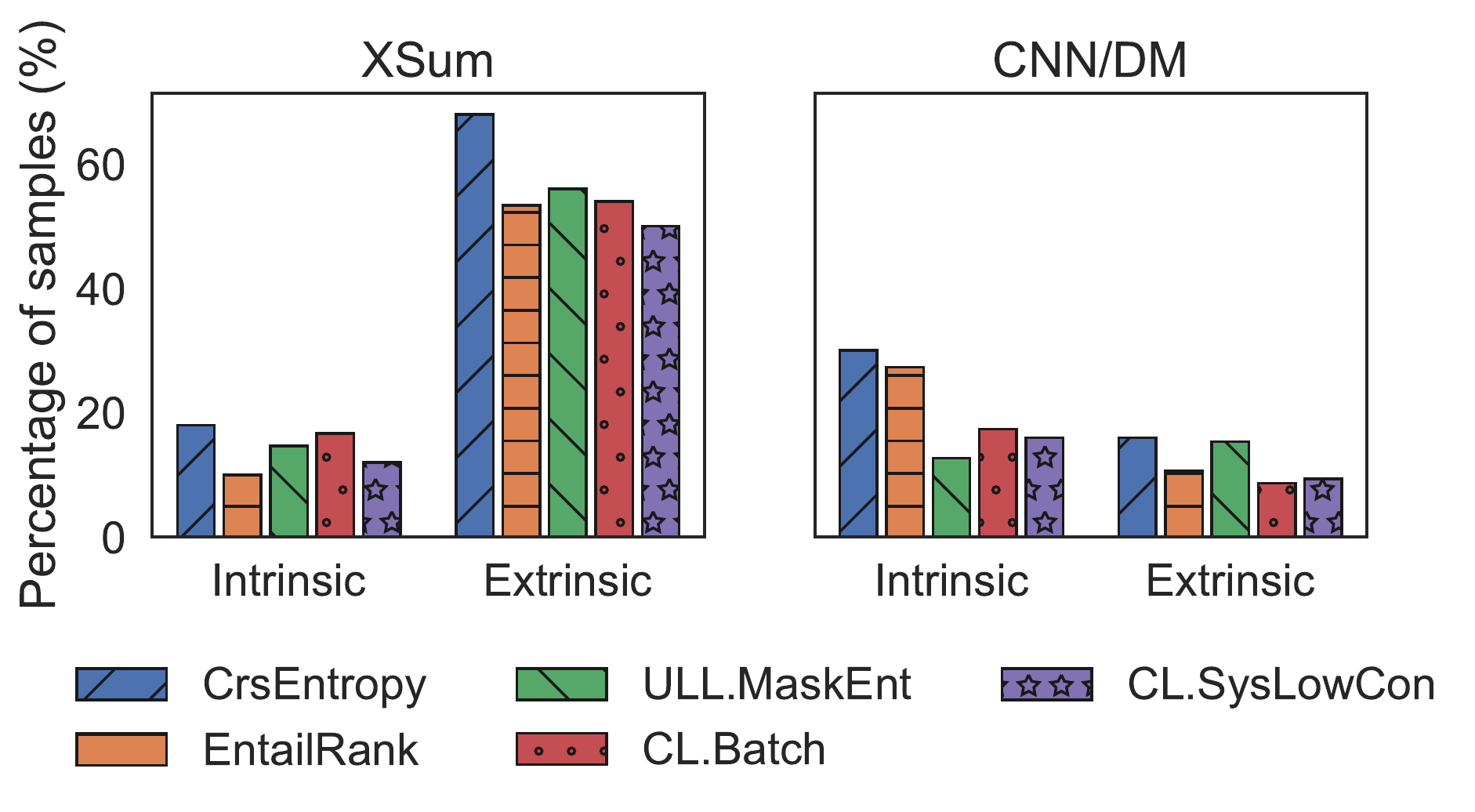}
    \vspace{-2mm}
    \caption{Portions of summaries with errors. CL models consistently reduce both types of errors.
    }
    \label{fig:human_eval_error_dist}
    \vspace{-3mm}
\end{figure}

\paragraph{Intrinsic vs. Extrinsic Errors.}
Next, the annotators are asked to label text spans with intrinsic and extrinsic errors as done in \S~\ref{sec:error_annotation}. 
Fig.~\ref{fig:human_eval_error_dist} shows that \textit{CL is more effective at reducing extrinsic errors than unlikelihood training can} on both datasets.
We also observe slight decreases of world knowledge in the summaries (figure attached in Appendix~\ref{appendix:human_eval}). 

\begin{figure}[t]
    \centering
    \includegraphics[width=0.47\textwidth]{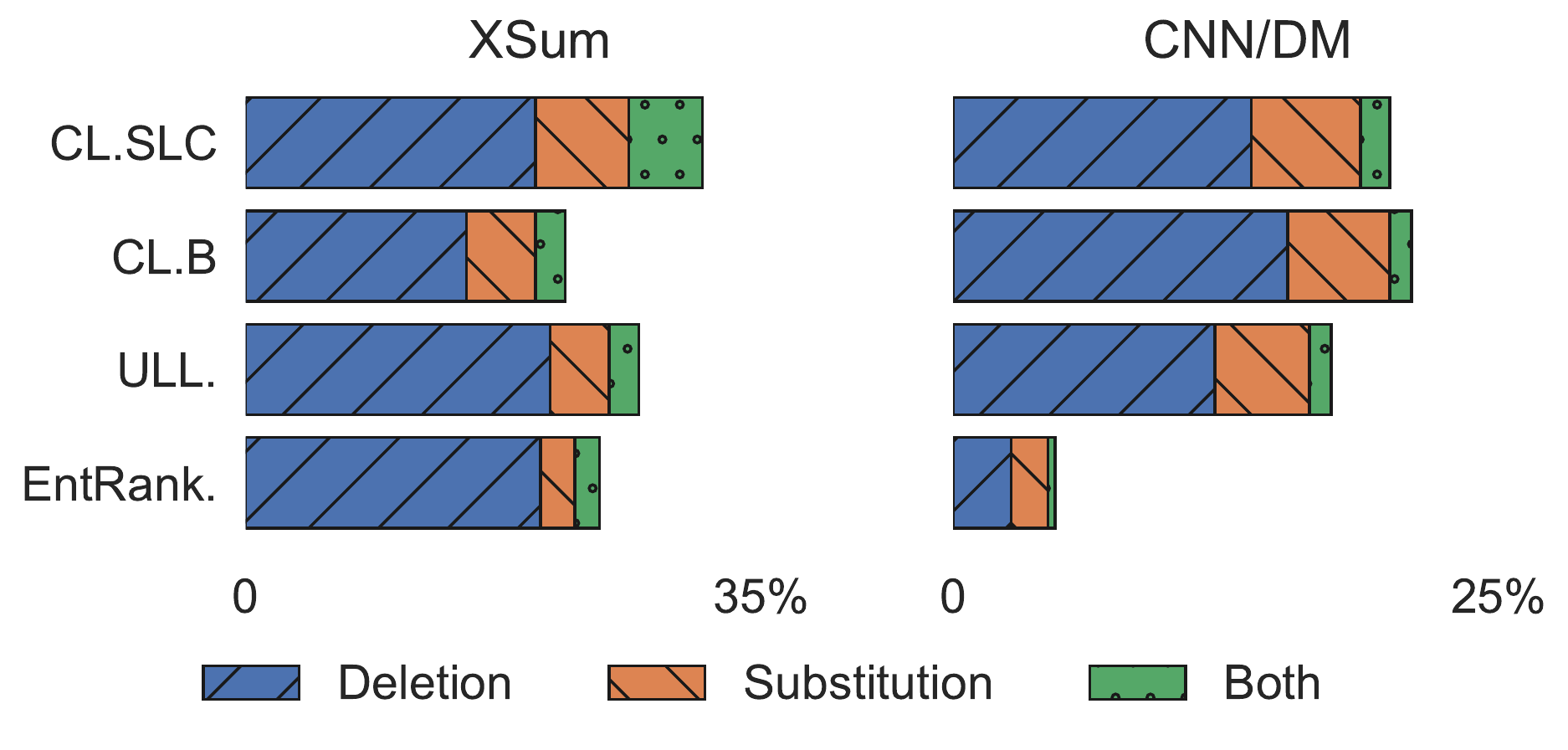}
    \vspace{-2mm}
    \caption{Summaries use different portions of error correction operations. 
    Contrastive learning with \textsc{SysLowCon} (CL.SLC) and \textsc{Batch} (CL.B) substitute errors with correct content more often than unlikelihood training with \textsc{MaskEnt} and \textsc{EntailRank}. 
    }
    \label{fig:error_correct_technique}
    \vspace{-3mm}
\end{figure}

\paragraph{Error Correction Operations.}
Finally, with reference to \textsc{CrsEntropy} summaries, human judges are instructed to label each system summary as whether it corrects any error by \textsc{CrsEntropy} using \textbf{deletion} of the incorrect content, \textbf{substitution} with factual information, or \textbf{both}. 
As seen in Fig.~\ref{fig:error_correct_technique}, CL-based models restore factually consistent information, e.g., by replacing erroneous names and numbers with correct ones, more frequently than unlikelihood training or entailment reranking.

\subsection{Variants of Summary Representation}
\label{subsec:ablaton}

Sample representation is critical for CL to be effective. Here we investigate summary representation variants as discussed in \S~\ref{sec:method}. 
There are two major considerations: (1) Should we consider all tokens in a summary or only representative ones (e.g., entities or last token)? (2) Should additional transformation, i.e., an MLP, be used?

Experiments on XSum using three negative sample construction strategies demonstrate that \textit{averaging the decoder outputs of all tokens and adding an MLP projection yield the best overall performance}, as shown in Table~\ref{tab:ablation_representation}. The implications are at least two-fold. 
First, even for entity- or relation-triggered sample modifications, using more global context helps with CL training. 
Second, additional transformation can help avoid model degeneration. For instance, more nonsensical and repetitive content is produced by variants without MLP.

\begin{table}[t]
    \centering
    \setlength{\tabcolsep}{2.5pt}
    \fontsize{8.5}{10}\selectfont
    \begin{tabular}{lccccccc}
    \toprule
        \multicolumn{2}{l}{} & \multicolumn{2}{c}{\textbf{\textsc{SwapEnt}}} & \multicolumn{2}{c}{\textbf{\textsc{MaskRel}}} & \multicolumn{2}{c}{\textbf{\textsc{SysLowCon}}} \\
        \textbf{Rep.} & \textbf{MLP} & \textbf{QEval} & \textbf{FC} & \textbf{QEval} & \textbf{FC} & \textbf{QEval} & \textbf{FC} \\
        \midrule
        \multicolumn{8}{l}{\textit{BART}} \\
        Last & \checkmark & 33.15 & 25.10 & 33.20 & 25.29 & 33.10 & 24.85 \\
        Last &  & \textcolor{red!53}{+0.13} & \textcolor{red!42}{+0.02} & \textcolor{blue!21}{--0.01} & \textcolor{blue!72}{--0.32} & \textcolor{blue!47}{--0.07} & \textcolor{blue!50}{--0.10} \\
        Entity & \checkmark & \hlc[xgreen]{\textbf{33.35}} & 25.41 & 33.34 & 25.44 & 33.32 & 24.46 \\
        Entity & & \textcolor{blue!50}{--0.13} & \textcolor{blue!47}{--0.07} & \textcolor{blue!41}{--0.14} & \textcolor{blue!45}{--0.05} & \textcolor{blue!69}{--0.29} & \textcolor{red}{+0.72} \\
        All & \checkmark & 33.30 & \hlc[xgreen]{\textbf{25.67}} & \hlc[xgreen]{\textbf{33.35}} & \hlc[xgreen]{\textbf{25.69}} & \hlc[xgreen]{\textbf{33.35}} & \hlc[xgreen]{\textbf{25.47}} \\
        All & & \textcolor{blue!43}{--0.23} & \textcolor{blue!90}{--0.80} & \textcolor{blue!44}{--0.04} & \textcolor{blue!88}{--0.48} & \textcolor{blue!46}{--0.26} & \textcolor{blue!80}{--0.40} \\
        \midrule
        \multicolumn{8}{l}{\textit{PEGASUS}} \\
        Last & \checkmark & 33.07 & \hlc[xgreen]{\textbf{25.45}} & 32.99 & 25.09 & 33.18 & 24.94 \\
        Last & & \textcolor{blue!47}{--0.07} & \textcolor{blue!96}{--0.56} & \textcolor{red!41}{+0.01} & \textcolor{blue!41}{-0.01} & \textcolor{blue!47}{--0.02} & \textcolor{blue!44}{--0.04} \\
        Entity & \checkmark & 33.03 & 25.43 & 33.05 & 24.77 & 33.20 & 24.59 \\
        Entity & & \textcolor{red!41}{+0.01} & \textcolor{blue!74}{--0.34} & \textcolor{blue!42}{--0.05} & \textcolor{red}{\textbf{\hlc[xgreen]{+0.64}}} & \textcolor{blue!70}{--0.30} & \textcolor{red!45}{+0.05} \\
        All & \checkmark & \hlc[xgreen]{\textbf{33.09}} & 25.09 & 33.06 & 25.28 & \hlc[xgreen]{\textbf{33.21}} & \hlc[xgreen]{\textbf{25.18}} \\
        All & & \textcolor{blue!51}{--0.11} & \textcolor{red!65}{+0.25} & \textcolor{red!43}{\hlc[xgreen]{\textbf{+0.03}}} & \textcolor{blue!59}{-0.19} & \textcolor{blue!42}{--0.02} & \textcolor{blue}{--0.80} \\
    \bottomrule
    \end{tabular}
    \vspace{-2mm}
    \caption{
    Comparing different formulations of summary representation in CL. 
    For models without MLP, we display score changes from their counterparts. 
    Overall, using all tokens with MLP produces better summaries. 
    }
    \label{tab:ablation_representation}
    \vspace{-2mm }
\end{table}

\section{Conclusion}
\label{sec:conclusion}

We present \textsc{CLIFF}, a contrastive learning-based framework to promote faithfulness and factuality of abstractive summaries. \textsc{CLIFF} uses both references and summaries that are factually inconsistent with the articles to train systems to be better at discriminating errors from factual and salient content. 
We further study strategies that automatically create erroneous summaries by editing from references or leveraging systems outputs, inspired by our new summary error analysis on state-of-the-art models. 
Both automatic evaluation and human ratings show that \textsc{CLIFF} achieves consistent improvements over competitive comparison methods, and is generalizable across datasets with systems fine-tuned from different large models.

\section*{Acknowledgements}
This research is supported in part by Oracle for Research Cloud Credits, National Science Foundation through a CAREER award IIS-2046016, and the Office of the Director of National Intelligence (ODNI), Intelligence Advanced Research Projects Activity (IARPA), via contract \# FA8650-17-C-9116. The views and conclusions contained herein are those of the authors and should not be interpreted as necessarily representing the official policies, either expressed or implied, of ODNI, IARPA, or the U.S. Government. The U.S. Government is authorized to reproduce and distribute reprints for governmental purposes notwithstanding any copyright annotation therein. 
We thank three anonymous reviewers for their valuable suggestions.

\bibliography{custom}
\bibliographystyle{acl_natbib}

\appendix


\section{Additional Analysis for Summary Error Annotation}
\label{appendix:error_annotation}

We hire two fluent English speakers to annotate summary errors on XSum  and CNN/DailyMail (CNN/DM). They annotate a common batch of 100 summaries generated by summarizers fine-tuned from BART and PEGASUS, with 50 articles in each batch. The two annotators are shown 50 HTML pages in a batch, each of which contains an article and two summaries generated by the two models. 
The detailed annotation guideline is given in Fig.~\ref{fig:annotation_guideline}.

\begin{figure}[t]
    \centering
    \includegraphics[width=0.45\textwidth]{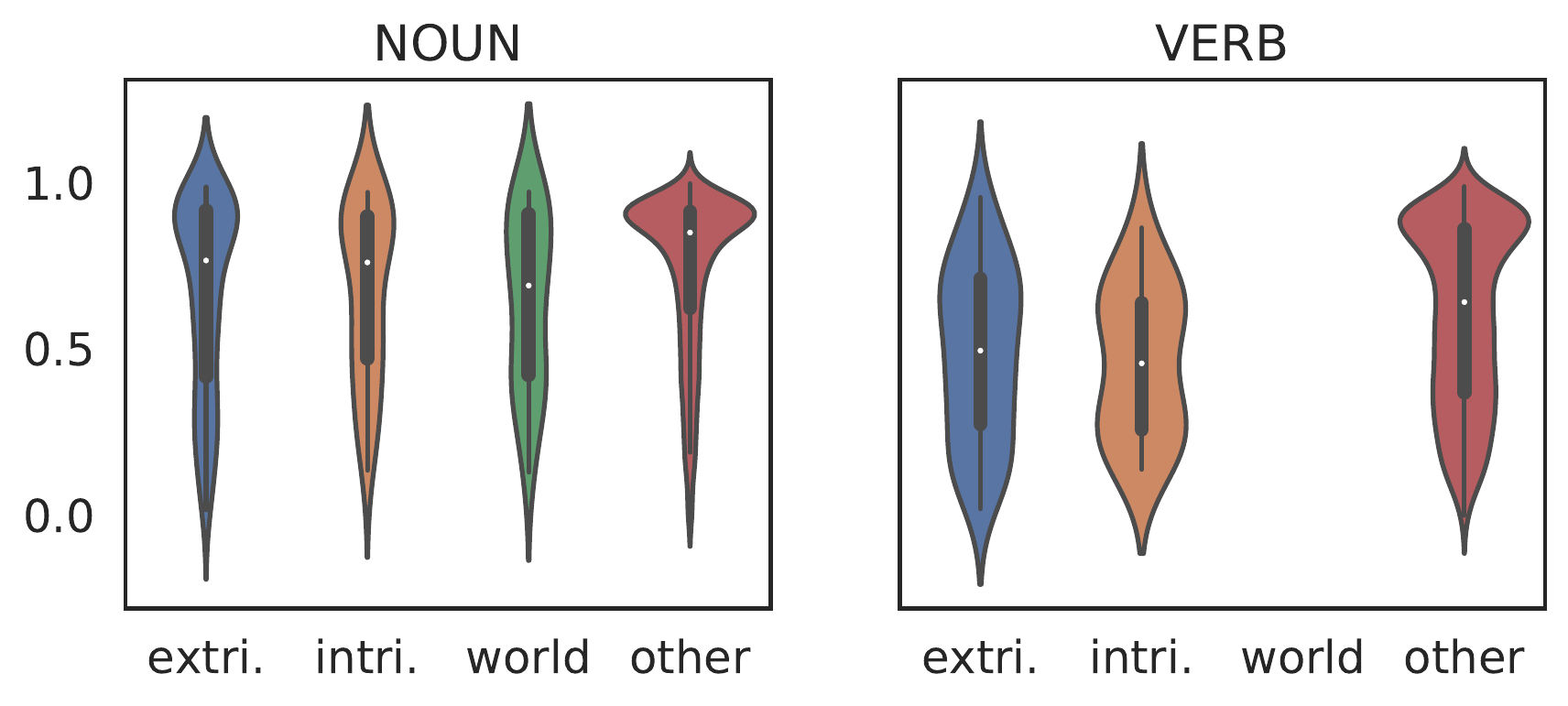}
    \caption{Probability distributions of generating the \textit{first} tokens of nouns and verbs, grouped by extrinsic errors, intrinsic errors, world knowledge, and other correct tokens. No verb is annotated as world knowledge.}
    \label{fig:first_token}
\end{figure}

\begin{figure}[t]
    \centering
    \includegraphics[width=0.45\textwidth]{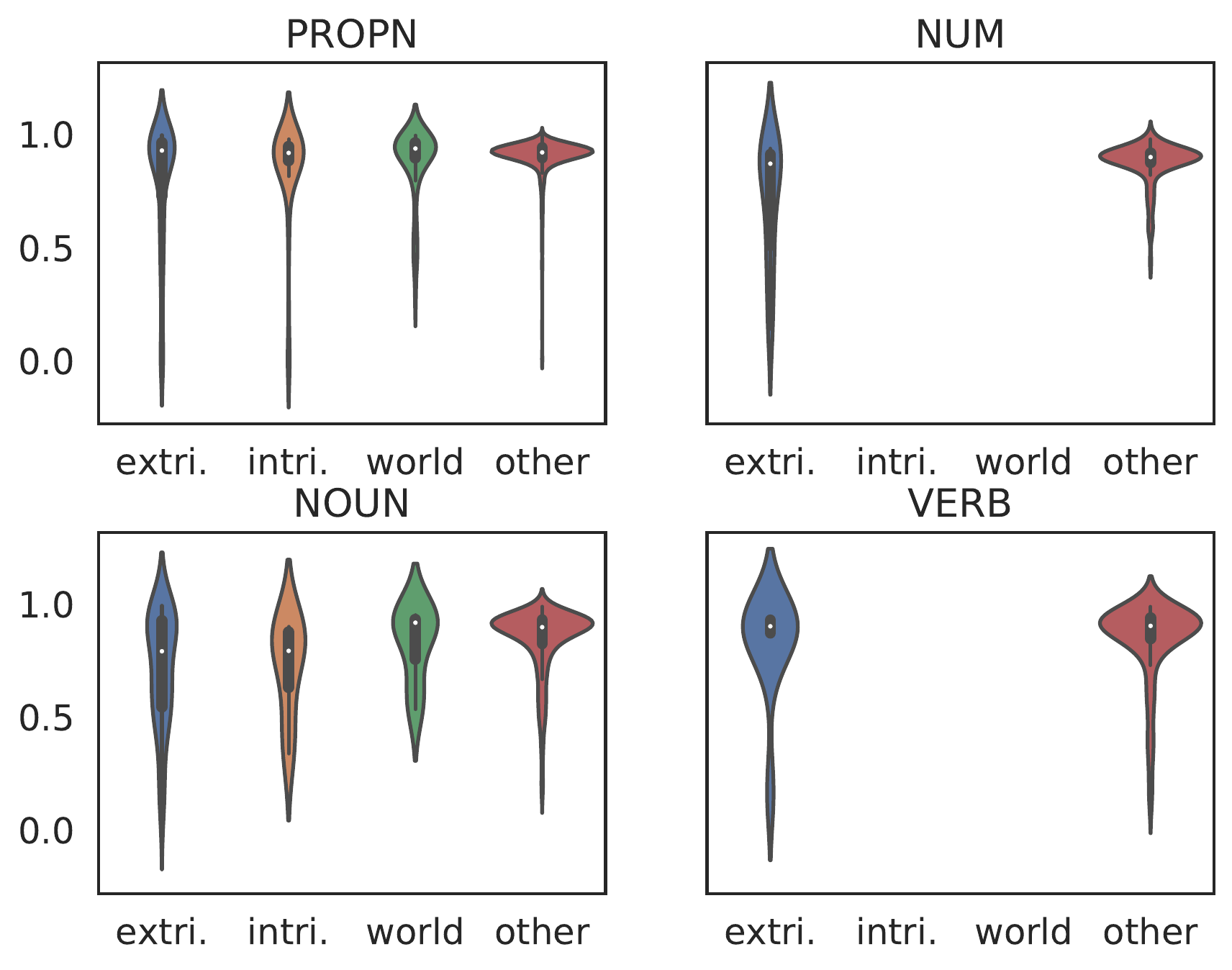}
    \caption{Probability distributions of generating the \textit{non-first} tokens of proper nouns, numbers, nouns, and verbs, grouped by extrinsic errors, intrinsic errors, world knowledge, and other correct tokens. Non-first tokens do not exist for numbers and verbs, as they only contain single tokens.}
    \label{fig:nonfirst_token}
\end{figure}

For our analysis on token generation probabilities, we additionally show the distributions of the first token's probability for nouns and verbs in Fig.~\ref{fig:first_token}. We also report the distributions of the non-first token's probability for proper nouns, numbers, nouns, and verbs in Fig.~\ref{fig:nonfirst_token}. As can be seen, tokens within extrinsic and intrinsic errors have high generation probabilities when they are non-first tokens.

\section{Statistics for Datasets and Training Samples}
\label{appendix:dataset_stat}

\paragraph{Summarization Datasets.}

We follow the official data splits for the two datasets, with the number of samples in each split listed in Table~\ref{tab:data_split}.

\begin{table}[t]
    \centering
    \small
    \begin{tabular}{lccc}
    \toprule
        \textbf{Dataset} & \textbf{Train} & \textbf{Validation} & \textbf{Test} \\
        \midrule
        XSum & 204{,}045 & 11{,}332 & 11{,}334 \\
        CNN/DM & 287{,}227 & 13{,}368 & 11{,}490 \\
        \bottomrule
    \end{tabular}
    \caption{Numbers of samples in train/validation/test splits of XSum and CNN/DM.}
    \label{tab:data_split}
\end{table}

\paragraph{Positive Samples.}

We observe unfaithful paraphrases by back-translation for some reference summaries, which are mainly due to the introduction of new entities and the rewriting of quoted text. Thus, we discard samples generated by back-translation that contain new entities and inconsistent quoted text. Finally, we obtain $182{,}114$ and $91{,}468$ positive samples by back-translation on XSum and CNN/DM.

\paragraph{Negative Samples.}

For consistency, we use the summarizer fine-tuned from BART in \textsc{RegenEnt}, \textsc{RegenRel} (\S~\ref{subsec:conditional}), and \textsc{SysLowCon} (\S~\ref{subsec:modelgeneration}) strategies. We tune a threshold to select negative samples from model generations in our \textsc{SysLowCon} strategy. The threshold is set to $0.21$, with F1 scores of $73.99$ and $40.49$ on XSum and CNN/DM annotated samples generated by BART.

The numbers of negative samples constructed by each strategy for training on XSum and CNN/DM are shown in Table~\ref{tab:num_neg_samples}.
\textsc{SysLowCon} constructs the least negative samples in total, while it achieves the best results as reported in our main paper (\S~\ref{subsec:autoeval}), indicating that its negative samples are more effective for training.

\begin{table}[ht]
    \centering
    \small
    \begin{tabular}{lcc}
    \toprule
        \textbf{Strategy} & \textbf{XSum} & \textbf{CNN/DM} \\
        \midrule
        \textsc{FCSample} & 936{,}164 & 1{,}291{,}710 \\
        \textsc{SwapEnt} & 438{,}003 & 1{,}617{,}764 \\
        \textsc{MaskEnt} & 360{,}795 & 1{,}050{,}200 \\
        \textsc{MaskRel} & 391{,}224 & 1{,}345{,}317 \\
        \textsc{RegenEnt} & 732{,}986 & 1{,}941{,}886 \\
        \textsc{RegenRel} & 993{,}694 & 1{,}453{,}044 \\
        \textsc{SysLowCon} & 401{,}112 & 502{,}768 \\
        \bottomrule
    \end{tabular}
    \caption{Numbers of negative samples constructed by different strategies on XSum and CNN/DM.}
    \label{tab:num_neg_samples}
\end{table}


\section{Implementation Details}
\label{appendix:implementation}

We use Fairseq~\cite{ott2019fairseq} and Huggingface Transformers~\cite{wolf-etal-2020-transformers} for our experiments with BART and PEGASUS. Our experiments are conducted on the RTX 8000 GPU with 48GB memory and the A100 GPU with 40GB memory.

\paragraph{Training Settings.}

For hyperparameters, we follow \citet{lewis-etal-2020-bart} for BART and \citet{zhang2020pegasus} for PEGASUS.
During training, we randomly select 5 and 4 negative samples for each input article in XSum and CNN/DM. Mixed-precision training is not supported by the PEGASUS implementation and is utilized on BART only.

\paragraph{Decoding Settings.}

We use the beam search algorithm to decode summaries.
For BART, we set the beam sizes as 6 and 4 on XSum and CNN/DM. A beam size of 8 is used for PEGASUS on both datasets.

\paragraph{Running Time and Model Sizes.}

The BART-based models take 6 and 13 hours for training on XSum and CNN/DM, and it takes 1.5 hour to decode on the two datasets.
Meanwhile, training the PEGASUS-based models takes 8 and 25 hours for XSum and CNN/DM, and the decoding takes 1 hour.

As for model sizes, our BART-based models and PEGASUS-based models have 400M and 568M parameters.


\section{Human Evaluation}
\label{appendix:human_eval}

\begin{figure}[t]
    \centering
    \includegraphics[width=0.45\textwidth]{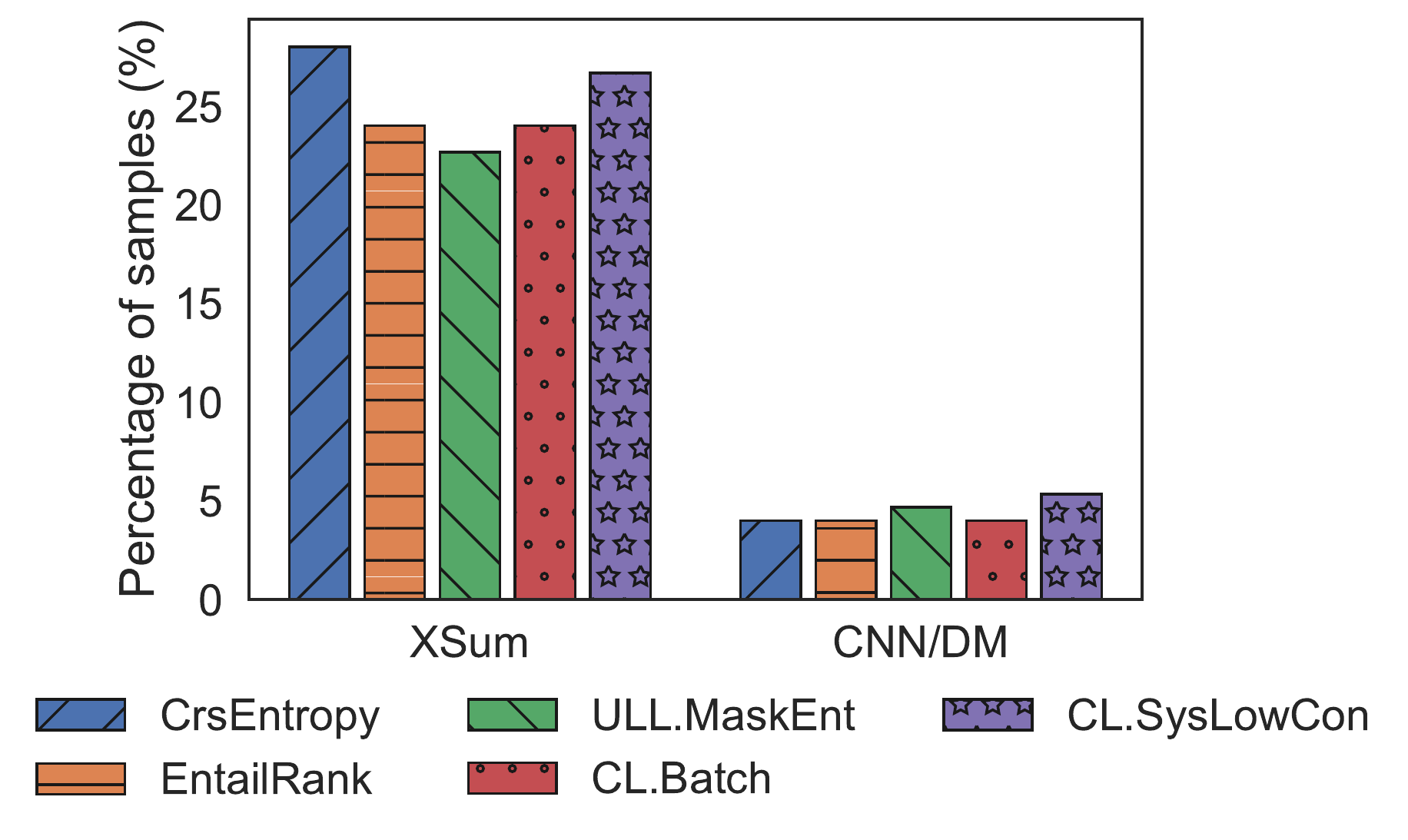}
    \caption{Percentages of samples containing world knowledge as labeled by human on the outputs of XSum and CNN/DM.}
    \label{fig:percent_world}
\end{figure}

In \S~\ref{subsec:humaneval}, we demonstrate the percentages of samples containing intrinsic errors and extrinsic errors for each model evaluated by human judges. 
Here, we report the percentages of samples containing world knowledge in Fig.~\ref{fig:percent_world}. On XSum, all models produce less world knowledge compared to the model trained with cross-entropy loss, while generating similar or greater amounts of samples with world knowledge on CNN/DM.

Our human evaluation guideline is shown in Fig.~\ref{fig:human_eval_guideline}.


\section{Sample Outputs}
\label{appendix:outputs}

We include more sample outputs in Fig.~\ref{fig:generation_examples}.


\begin{figure*}[t]
    \centering
    \fontsize{10}{12}\selectfont
    \begin{tabular}{p{0.88\textwidth}}
    \toprule
        In this study, you will first read article-summary pairs and then identify three types of text spans in the summaries. These spans include content that is contradicted by or cannot be implied from the article. The description for each type is described below: \\
        \begin{itemize}
            \item \textbf{Intrinsic errors:} Text spans that misconstruct phrases or clauses from the article. 
            \item \textbf{Extrinsic errors:} Text spans that include words that are not in the article and are not verifiable or cannot be verified by Wikipedia. 
            \item \textbf{world knowledge:} Text spans that contain information that is not covered by the article but can be validated by Wikipedia. 
        \end{itemize} \\
        When selecting spans, you should always make sure the spans are complete words. \\
        \\
        In practice, you should follow the these steps carefully: (1) read the article and summaries carefully; (2) figure out if there is content contradicted by or not presented in the article; (3) label the span as an intrinsic error if it misconstructs phrases or clauses from the article; (4) if the span does not belong to intrinsic errors, search within Wikipedia and determine whether the content in the span can be verified; (5) label it as world knowledge if the it can be verified by Wikipedia, otherwise label it as an extrinsic error. \\
        \midrule
        \textbf{Example annotations 1} \\
        \textbf{Article:} Isis Academy in Oxford said it had rebranded as ``Iffley Academy'' to protect its ``reputation, integrity and image''. \textbf{The name `Isis' was originally chosen as the school is near to the section of the River Thames of the same name}. Formerly Iffley Mead School, it became Isis Academy in 2013. A statement issued by the school said it had changed name following ``the unforeseen rise of ISIS (also known as ISIL and the Islamic State) and related global media coverage of the activities of the group''. ``Our priority is to remove the detrimental impact which the name `Isis' had on pupils, their families and our staff.'' Last year a language school in the city removed Isis from its name for the same reason. The Isis is the name given to the part of the River Thames above Iffley Lock in Oxford. It is also the name of the goddess wife of the god Osiris in Egyptian beliefs. \\
        
        \textbf{Summary:} A school that \hlc[crimsonglory!40]{was named after} the Islamic State (IS) militant group has changed its name. \\
        \textbf{Explanation:} \textit{``was name after''} is an intrinsic error contradicted by the article sentence in \textbf{bold}. \\
        \midrule
        \textbf{Example annotations 2} \\
        \textbf{Article:} Khalil Dale, 60, was abducted in \textbf{Quetta} in January 2012 and was found dead on a roadside a few months later. He had been beheaded. A note next to his body said he was killed because a ransom had not been paid. Mr Dale was born in York but lived in Dumfries. He spent 30 years working in countries including Somalia, Afghanistan and Iraq. An inquest into his death was held at Chesterfield Coroners Court because he is buried in Derbyshire. The court heard that the Muslim convert, who was formerly known as Kenneth, worked as a humanitarian assistance relief worker. Following his abduction, negotiations were undertaken by the International Committee of the Red Cross with the help of the UK government. His body was found on 29 April 2012. The inquest was told that he died as a result of decapitation. Senior coroner Dr Robert Hunter concluded that Mr Dale was unlawfully killed while providing international humanitarian assistance. \\
        \textbf{Summary:} A British aid worker was unlawfully killed by \hlc[bleudefrance!40]{Islamist militants} in \hlc[yellow!60]{Pakistan}, an inquest has heard. \\
        \textbf{Explanation:} \textit{``Islamist militant''} is an extrinsic error as it can not be found in or inferred from the article. The information is also not verifiable by Wikipedia. \textit{``Pakistan''} is world knowledge as \textbf{Quetta} in the article is a city in Pakistan according to Wikipedia. \\
        \bottomrule
    \end{tabular}
    \caption{Guideline for our summary error annotation (\S~\ref{sec:error_annotation}).}
    \label{fig:annotation_guideline}
\end{figure*}

\begin{figure*}[t]
    \centering
    \fontsize{10}{12}\selectfont
    \begin{tabular}{p{0.92\textwidth}}
    \toprule
        In this study, you will evaluate 100 sets of summaries produced by four systems. For each set, its corresponding article and a baseline summary are shown before the four system summaries. The errors in the baseline summary are highlighted. \\
        Please \textit{first read the article and the baseline summary} and then \textit{compare each system summary against the baseline summary} based on \textbf{informativeness} and \textbf{factual consistency}. In addition, please decide the \textbf{operations} made by the system to achieve better factual consistency. \\
        For informativeness and factual consistency, you need to label whether the system summary is better or worse than the baseline summary. You can also label the system summary as tying with the baseline summary. \\
        You need to consider two types of operations: \textbf{deletions} and \textbf{substitutions}. Please label the system summary as making deletions, substitutions, or both operations. Examples for the aspects and the operations are as follows. \\
        \midrule
        
        \textbf{Article:} Alexys Brown, also known as Lexi, died at her home in Emmadale Close, Weymouth, on Thursday. An investigation is under way to discover how she became trapped. A post-mortem examination is due to be carried out this week. It was originally hoped the appeal would raise £2,000. Alison Record, who started the Just Giving appeal, said she was "heart broken" over the death. ``Everybody by now has heard of the terrible tragedy the Brown family have suffered with the loss of their beautiful and beloved little girl Lexi,'' the appeal page reads.  Many other comments have been posted on the appeal page. Steph Harris said: ``Thinking of you all at this devastating time, fly high beautiful princess. Love Steph and family xxx'' Lesley Andrews added: ``No amount of money will take away the pain, but so much love comes with every penny. Take care. xx'' Aster Group, the housing association responsible for managing the home, is assisting with the police investigation. The Health and Safety Executive (HSE) is also investigating. Dorset County Council said it had not installed the disabled lift at the property. \\
        
        \textbf{Baseline Summary:} An appeal to raise \hlc[crimsonglory!40]{10,000 pounds} for the family of a \hlc[bleudefrance!40]{three-year-old} girl who died after becoming trapped in a lift has raised \hlc[bleudefrance!40]{more than 20,000 pounds}. \\
        
        \smallskip
        \textbf{Informativeness:} Whether the summary captures salient content from the input article. Note that incorrect content should be considered as invalid. \\
        \textbf{Win.}  \textit{An appeal to raise money for the family of a three-year-old girl who died after getting stuck in a lift was originally hoped for raising £2,000.} The target money of the appeal is a salient information. \\
        \textbf{Tie.}  \textit{An appeal to raise money for the family of a girl who died after getting stuck in a lift has raised more than £20,000.} Compared to the baseline, missing incorrect information does not affect the informativeness. \\
        \textbf{Lose.} \textit{An appeal to raise money for the family of a three-year-old girl has raised more than £20,000.} This system summary does not mention the death of the girl, which is a salient content of the article. \\
        
        \smallskip
        \textbf{Factual Consistency:} Whether the summary is factually correctly based on the article and knowledge from Wikipedia. \\
        \textbf{Win.}  \textit{An appeal has been set up for the family of an \hlc[bleudefrance!40]{eight-year-old} girl who died after becoming trapped in a lift at her Dorset home.} This system summary does not generate the incorrect numbers of money. \\
        \textbf{Tie.} \textit{An appeal to raise \hlc[crimsonglory!40]{5,000 pounds} for the family of a \hlc[bleudefrance!40]{seven-year-old} girl who died after becoming trapped in a lift has raised \hlc[bleudefrance!40]{more than 20,000 pounds}.} This system summary makes similar errors to the baseline. \\
        \textbf{Lose.} \textit{\hlc[crimsonglory!40]{The family} of an \hlc[bleudefrance!40]{eight-year-old} girl who died after becoming trapped in a lift at her Dorset home \hlc[crimsonglory!40]{have set a fundraising target} of \hlc[bleudefrance!40]{10,000 pounds}.} This system summary fabricates an event \textit{The family have set a fundraising target}, which is more severe than errors of modifiers. \\
        
        \smallskip
        \textbf{Deletion:} The incorrect content in the baseline summary is deleted. \\
        - \textit{An appeal for the family of a \hlc[bleudefrance!40]{three-year-old} girl who died after becoming trapped in a lift has raised \hlc[bleudefrance!40]{more than 20,000 pounds}.} The error \textit{``10{,}000 pounds''} is deleted. \\
        
        \textbf{Substitution:} The incorrect content in the baseline summary is replaced with correct one. \\
        - \textit{An appeal to raise 2,000 pounds for the family of a \hlc[bleudefrance!40]{three-year-old} girl who died after becoming trapped in a lift has raised \hlc[bleudefrance!40]{more than 20,000 pounds}.} The error \textit{``10{,}000 pounds''} is substituted with \textit{``2,000 pounds''}, which is the correct information. \\
        
        \bottomrule
    \end{tabular}
    \caption{Guideline for our human evaluation (\S~\ref{subsec:humaneval}).}
    \label{fig:human_eval_guideline}
\end{figure*}

\begin{figure*}[th]
    \centering
    \small
    \begin{tabular}{p{0.95\textwidth}}
    \toprule
    \textbf{Example 1} \\
    \midrule
        \textbf{CNN/DM Article:} At the grand old age of 75, Jack Nicklaus is still capable of hitting aces. The Golden Bear added another magic moment to his storied career at Augusta National in the Par-3 Contest. Stepping up to the tee on the 130-yard fourth, the greatest golfer of all time saw his shot sail beyond the flag before spinning back into the hole. Jack Nicklaus gave the crowd something to cheer with a hole in one on the fourth during the Par-3 Contest. Nicklaus holds up his ball to an adoring crowd as Gary Player (left) and Ben Crenshaw salute the great. Crenshaw and Nicklaus fist pump following his ace on the 130-yard hole at Augusta National. Nicklaus was playing alongside Gary Player and Ben Crenshaw as the first roar of Masters week filled the skies in a sunswept Georgia. 'I had an interview this morning, and I said "well, all I've got to do is go out and win the Par-3 and make a hole and one", and I make a hole in one,' Nicklaus said afterwards. . The ace was Nicklaus' first on the hallowed turf of Augusta, but the veteran saw his achievement eclipsed by that of Camilo Villegas on a day which saw five hole-in-ones, levelling the record from 2002. The Colombian followed Nicklaus with an ace on the fourth and also managed one on the eighth, but he couldn't quite land the title, losing out to world No 74 Kevin Streelman in a play-off. Nicklaus became the youngest person to wear a green jacket in 1963, and collected his sixth in 1986. He is one of five men to complete the career grand slam, an accolade which favourite Rory McIlroy can achieve if he wins his third major in succession.  \\
        \midrule
        \textbf{Cross-entropy:} Jack Nicklaus hit a hole in one on the fourth during the Par-3 Contest. The 75-year-old was playing alongside Gary Player and Ben Crenshaw. Camilo Villegas and \hlc[crimsonglory!40]{Kevin Streelman} also hit hole-in-ones on the day. \\
        \textbf{Entailment reranking:} Jack Nicklaus hit a hole in one on the fourth during the Par-3 Contest. The 75-year-old was playing alongside Gary Player and Ben Crenshaw. Camilo Villegas and \hlc[crimsonglory!40]{Kevin Streelman} also hit hole-in-ones on day one. \\
        \textbf{Unlikelihood training:} Jack Nicklaus hit a hole in one on the fourth during the Par-3 Contest. The 75-year-old was playing alongside Gary Player and Ben Crenshaw. Camilo Villegas followed Nicklaus with a hole-in-one on the same hole. Nicklaus is one of five men to complete the career grand slam.  \\
        \textbf{Contrastive learning:} Jack Nicklaus hit a hole in one on the fourth during the Par-3 Contest. The 75-year-old was playing alongside Gary Player and Ben Crenshaw. Camilo Villegas followed Nicklaus with a hole-in-one on the eighth. \\
    \midrule
    \textbf{Example 2} \\
    \midrule
        \textbf{XSum Article:} The British supermarket giant, the world's third largest retailer, has been looking to dispose of its loss-making US food chain for some time. Around 4,000 of Fresh \& Easy's 5,000 employees are to transfer to the new business as part of the deal. Tesco said in a statement that this is an "orderly and efficient exit from the US market". Under the terms of the deal, Tesco will loan the new business around £80m. Fresh \& Easy, which operates 200 stores across California, Nevada and Arizona, has lost money since opening in 2007. Tesco did not specify which stores are being transferred to Yucaipa, but it is understood the remaining 50 shops will be closed. It will cost Tesco £150m to dispose of the Fresh \& Easy brand, but part of that is a loan which may be paid back. Also, if Yucaipa is sucessful and the business becomes profitable, Tesco has the option to buy a stake. Yucaipa is a Los Angeles based private equity company focused on supermaket and restaurant businesses and is run by billionaire investor Ron Burkle. In April of this year, Tesco reported its first fall in annual profits for 20 years.  \\
        \midrule
        \textbf{Cross-entropy:} Tesco has agreed to sell its US food chain Fresh \& Easy to Yucaipa in a deal \hlc[bleudefrance!40]{worth about 150m}, the BBC understands. \\
        \textbf{Entailment reranking:} Tesco is to sell its US food chain Fresh \& Easy to Yucaipa \hlc[bleudefrance!40]{for an undisclosed sum}. \\
        \textbf{Unlikelihood training:} Tesco has agreed to sell its US food business, Fresh \& Easy, to a private equity firm called Yucaipa \hlc[bleudefrance!40]{for an undisclosed sum}.  \\
        \textbf{Contrastive learning:} Tesco has agreed to sell its US food chain, Fresh \& Easy, to Yucaipa. \\
        \midrule
        \textbf{Example 3} \\
    \midrule
        \textbf{XSum Article:} Neil Aspin's promotion-chasing hosts have not lost in nine National League matches while Adam Lockwood's side are unbeaten in five. Guiseley went ahead on 15 minutes against the run of play when a throw-in found James Hurst who squared to Jake Lawlor to stroke into an empty net. Gateshead defender Liam Hogan superbly blocked Jordan Preston's effort and Guiseley keeper Jonny Maxted then saved well from Wesley York's shot just before the break. The hosts, who started the second half well, levelled on 62 minutes when a slip by half-time substitute Derek Asamoah let York curl sweetly into the top-right corner from the edge of the box. Match report supplied by the Press Association. Match ends, Gateshead 1, Guiseley 1. Second Half ends, Gateshead 1, Guiseley 1. Substitution, Guiseley. Michael Rankine replaces Jordan Preston. Substitution, Gateshead. Luke Hannant replaces Gus Mafuta. Gus Mafuta (Gateshead) is shown the yellow card. Substitution, Guiseley. Adam Boyes replaces Jake Cassidy. Goal! Gateshead 1, Guiseley 1. Wes York (Gateshead). Substitution, Guiseley. Derek Asamoah replaces Kevan Hurst. Second Half begins Gateshead 0, Guiseley 1. First Half ends, Gateshead 0, Guiseley 1. Simon Walton (Guiseley) is shown the yellow card. Goal! Gateshead 0, Guiseley 1. Jake Lawlor (Guiseley). First Half begins. Lineups are announced and players are warming up.  \\
        \midrule
        \textbf{Cross-entropy:} Gateshead and Guiseley maintained their \hlc[crimsonglory!40]{unbeaten starts to the season} with a 1-1 draw at the International Stadium. \\
        \textbf{Entailment reranking:} Gateshead and Guiseley shared the spoils after a 1-1 draw at the International Stadium. \\
        \textbf{Unlikelihood training:} Gateshead and Guiseley shared the spoils after a \hlc[crimsonglory!40]{goalless} draw in the National League.  \\
        \textbf{Contrastive learning:} Gateshead and Guiseley shared the spoils after a 1-1 draw at Gateshead. \\
     \bottomrule
        
    \end{tabular}
    \caption{Sample generated summaries by fine-tuned BART models. Intrinsic errors are highlighted in \hlc[crimsonglory!40]{red} and extrinsic errors are in \hlc[bleudefrance!40]{blue}. 
    \textsc{MaskEnt} and \textsc{SysLowCon} are used for negative sample construction with unlikelihood training and contrastive learning.}
    \label{fig:generation_examples}
\end{figure*}



\end{document}